\documentclass[10pt,twocolumn,letterpaper]{article}

\usepackage[pagenumbers]{iccv}      

\usepackage{xspace}
\usepackage{graphicx}
\usepackage{url}
\usepackage{float}
\usepackage{wrapfig}
\usepackage{wrapfig}
\usepackage{caption}
\usepackage{pifont}
\usepackage{url}
\usepackage{amsmath}
\usepackage{bbm} 
\usepackage{multirow}
\usepackage{graphicx}
\usepackage{colortbl}
\usepackage{adjustbox}
\usepackage{subcaption}
\usepackage{booktabs}
\usepackage{xcolor}

\newcommand{\Model}{DiagNote}
\newcommand{\Dataset}{MMDiag}

\newcommand{\Agent}{Deliberate}
\newcommand{\Grounder}{Gaze}

\definecolor{iccvblue}{rgb}{0.21,0.49,0.74}
\usepackage[pagebackref,breaklinks,colorlinks,allcolors=iccvblue]{hyperref}

\title{Taking Notes Brings Focus? Towards Multi-Turn Multimodal Dialogue Learning}

\author{
Jiazheng Liu\textsuperscript{\rm 1}\thanks{Work done as an intern at BAAI.} \ \  Sipeng Zheng\textsuperscript{\rm 2} \ \   Börje F. Karlsson\textsuperscript{\rm 2} \ \  Zongqing Lu\textsuperscript{\rm 1,2}\thanks{Correspondence to $<$zongqing.lu@pku.edu.cn$>$}\\ \\
\textsuperscript{\rm 1}School of Computer Science, Peking University \\
\textsuperscript{\rm 2}Beijing Academy of Artificial Intelligence  \\
}

\begin{document}

\hyphenpenalty=4000
\tolerance=100000

\maketitle

\begin{abstract}
Multimodal large language models (MLLMs), built on large-scale pre-trained vision towers and language models, have shown great capabilities in multimodal understanding. However, most existing MLLMs are trained on single-turn vision question-answering tasks, which do not accurately reflect real-world human conversations. In this paper, we introduce \Dataset, a multi-turn multimodal dialogue dataset. This dataset is collaboratively generated through deliberately designed rules and GPT assistance, featuring strong correlations between questions, between questions and images, and among different image regions; thus aligning more closely with real-world scenarios. \Dataset\ serves as a strong benchmark for multi-turn multimodal dialogue learning and brings more challenges to the grounding and reasoning capabilities of MLLMs. Further, inspired by human vision processing, we present \Model, an MLLM equipped with multimodal grounding and reasoning capabilities. \Model\ consists of two modules (\Agent\ and \Grounder) interacting with each other to perform Chain-of-Thought and annotations respectively, throughout multi-turn dialogues. We empirically demonstrate the advantages of \Model\ in both grounding and jointly processing and reasoning with vision and language information over existing MLLMs. 
\end{abstract}
\section{Introduction}
\begin{figure}
\centering
  \includegraphics[width=0.9\linewidth]{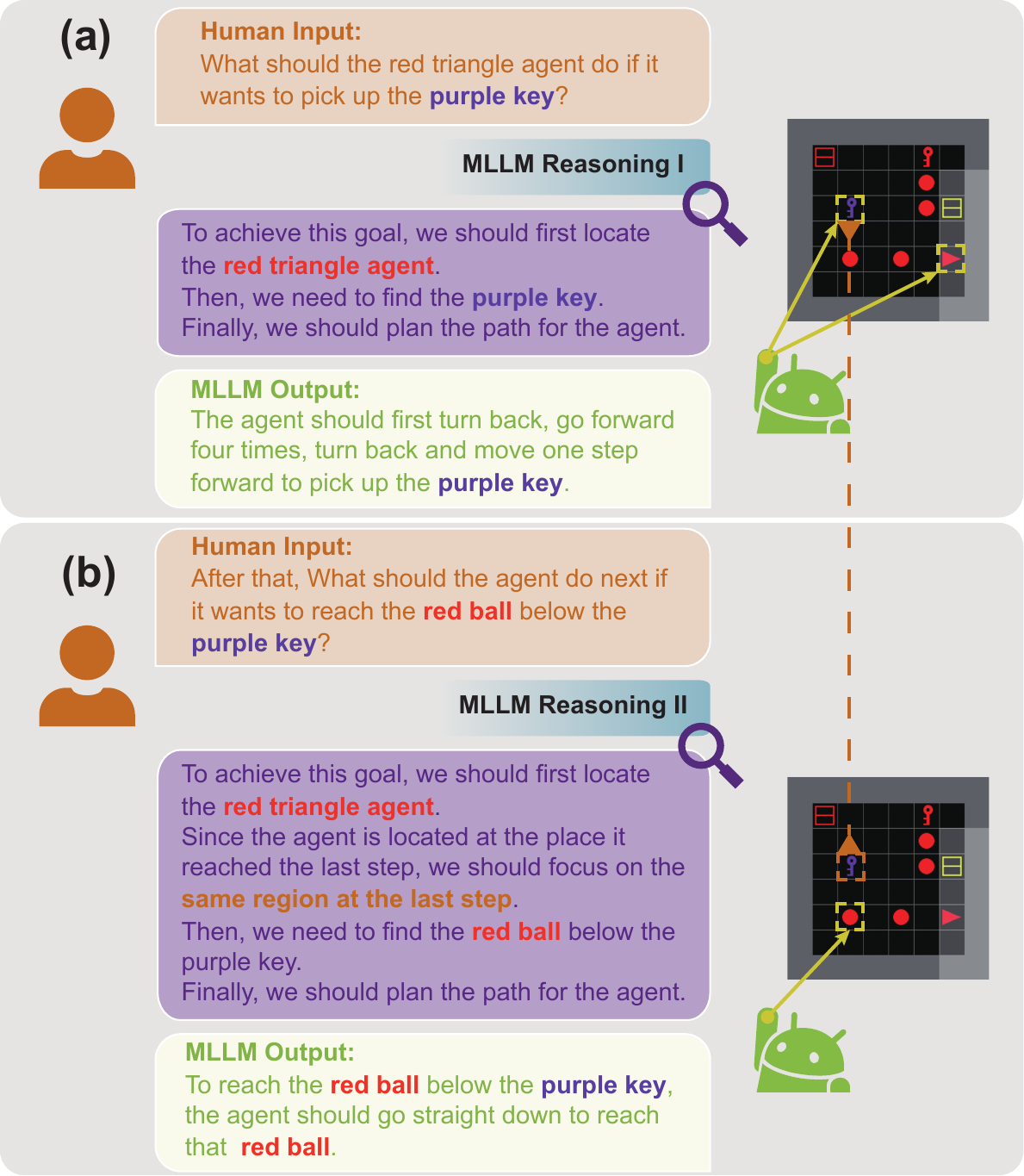}
   \vspace{-5pt}
   \captionof{figure}{Multi-turn multimodal dialogue: \textbf{(a) Saliency tracking.} The MLLM needs to focus on both the red triangle agent and the purple key, which scatter on the image, to answer the question correctly. \textbf{(b) Saliency recall. } The MLLM needs to retain focus on the region where the agent will stop after the last question. 
   }
   \label{fig:sketch}
   \addvspace{-10pt}
\end{figure}

In recent years, large language models (LLMs) have achieved remarkable advances in various natural language applications, including chatbots~\cite{qwen,gpt4report,gemini}, programming assistants~\cite{cursor}, and rhetorical aides~\cite{deepl}.
The success has further spurred the development of multimodal large language models (MLLM)~\cite{llava,zheng2025unicode}.
However, most existing MLLMs are trained as single black-box systems to handle multimodal instructions, often struggling with inaccuracies and hallucinations, especially in complex multi-turn dialogues~\cite{tan2024towards,steve-eye}.
We hypothesize such challenges arise from the MLLM's difficulty in maintaining focus on target regions throughout the conversation, especially for high-resolution images with overly long visual tokens.
In this paper, we seek to address these issues by moving beyond a black-box approach to an explicit target-grounding solution.
Here, we summarize two key goals for multi-turn multimodal dialogue learning:
\ding{182} ``saliency tracking'', where the MLLM must keep tracking different relevant regions over the course of the dialogue, 
and \ding{183} ``saliency recall'', where the model needs to consistently retain focus on the same critical information across multiple question-answering (QA) rounds.
For example, in the dialogue illustrated in \Cref{fig:sketch}, completing the Minigrid~\cite{Minigrid} task requires the MLLM to accurately locate both the agent (\ie ``\textcolor{red}{red triangle}'') and the target (\ie ``\textcolor{purple}{purple key}'') to answer the initial question.
The following question then builds upon this information, requiring the MLLM to reason about the agent's starting position based on the previously identified location of the key.  
This example illustrates the need for sustained and explicit grounding to multiple specific visual details in multi-turn multimodal dialogue.

To achieve these two goals, we draw inspiration from how humans maintain focus while studying. 
For instance, when working through documents, people may lose concentration, but can quickly refocus by using simple techniques such as jotting down notes or highlighting key points. 
Even basic marks, such as circling or underlining, can significantly enhance focus without requiring elaborate explanations.
These visual cues guide attention, making it easier to track, recall, and revisit important information.
In contrast, existing MLLMs lack such tracking capabilities, prompting us to ask: ``Can an MLLM be designed to equip similar attention-guiding abilities? If so, what would that model design entail?''

To answer this question, we first review existing tuning methods for MLLMs and identify a critical gap: the lack of quality multi-turn multimodal dialogue datasets that adequately reason over both visual and text information.
Existing datasets, such as MMDU~\cite{mmdu} and SciGraphQA~\cite{scigraphqa}, primarily consist of single-turn QA pairs, where most questions can be answered independently without relying on prior context.
To bridge this gap, we introduce a novel dataset, \Dataset, designed as a foundational benchmark for challenging multi-turn multimodal dialogue.
This dataset offers visually detailed multi-turn dialogues across a range of scenarios.

Furthermore, recent studies have introduced various modules to help keep focus in multi-turn multimodal dialogues.
However, these methods either ``zoom in'' to progressively narrow focus areas with the aid of external grounding and OCR tools~\cite{cogcom}, or identify a single region of interest per question before generating an answer~\cite{viscot}.
These approaches lead to severe limitations: the zoom-in method restricts the focus to smaller regions, potentially missing broader context, while the single-region method isolates specific areas, overlooking multiple relevant details that could enrich responses.
To address these limitations, we propose \Model, a model designed to enhance focus and reasoning in multi-turn multimodal dialogue.
\Model\ comprises two main modules: \Agent\ and \Grounder.
The \Agent\ module guides the \Grounder\ module in dynamically adjusting the region of visual focus, while the \Grounder\ module highlights crucial areas for subsequent processing by the \Agent\ module. These two modules interact across multiple dialogue turns, emulating human visual processing to produce an answer accompanied by optional reasoning and grounding steps.
Through this interactive mechanism, \Model\ can achieve more effective reasoning with multimodal information, resulting in accurate and context-aware responses throughout multi-turn dialogues.

Our main contributions are summarized as follows:
\ding{182} To address the need for robust multimodal grounding and reasoning, we build a new large-scale multi-turn multimodal dialogue dataset -- \Dataset\ -- across several QA scenarios (\eg daily life and tabular data), using rule-based searching and GPT-4o-mini~\cite{gpt4omini} capabilities.
\ding{183} Inspired by human visual processing, we propose \Model\ and its two key modules -- \Agent\ and \Grounder\ -- to enhance the model’s capacity for multimodal information integration and reasoning. 
\ding{184} We evaluate \Model's reasoning and grounding abilities on \Dataset\ and other benchmarks and the results demonstrate that the introduction of \Dataset\ and \Model\ significantly improves performance in multimodal conversations, while the \Dataset\ itself can also serve as a more challenging benchmark for this area.
\vspace{-10pt}
\section{Related Work}

\subsection{Multimodal Large Language Models}

The introduction of Transformers~\cite{transformer,swin_transformer} and large-scale training has greatly enhanced model capabilities, leading to the development of advanced vision encoders~\cite{clip} and large language models (LLMs)~\cite{vicuna, llama}. 
Building on these advancements, multimodal large language models (MLLMs)~\cite{llava, steve-eye} have demonstrated impressive performance across a wide range of multimodal tasks, and potential applications from VR/AR to game agents~\cite{game-mllm-survey, feng2024llama}.

An MLLM typically consists of three main components: modality encoders, modality interfaces, and LLMs~\cite{mllmsurvey}. Modality encoders and LLMs process modality information and language separately, and then modality interfaces align other modalities with the representations of the language. 
For modality interfaces, most approaches~\cite{llava, blip2} rely on learnable connectors. For modality encoders, research indicates that visual information processing (especially in terms of image resolution~\cite{mm1}) significantly affects the performance of MLLM. 
Additionally, certain models incorporate generators to produce other multimodal data, such as low-level actions~\cite{palme} or images~\cite{steve-eye}.

MLLM training commonly follows a two-stage process. In the first stage, vision and language modalities are aligned with the modality interface, often through pre-training on large datasets of image-caption pairs~\cite{llava, laion5b, cc12m}. The second stage involves fine-tuning with visual question-answering (VQA) tasks~\cite{llava,TextVQA} for better LLMs' capabilities of instruction following. This two-stage process is widely used in MLLMs like PALI-X~\cite{palix}, Qwen-VL~\cite{qwenvl}, and LLaVA~\cite{llava}, forming a strong foundation for subsequent MLLM advancements.

\subsection{Grounding and Reasoning Benefit MLLMs}

MLLMs can perform in-context learning~\cite{ICL}, enabling generalization to new tasks from a few examples. The Chain-of-Thought (CoT)~\cite{cot} reasoning mechanism also allows models to approach problem-solving step-by-step. However, when faced with unfamiliar tasks, MLLMs sometimes rely excessively on the generalization capabilities of the LLM component, leading to overlooking visual details and hallucinations.
To address these limitations, models like CogCoM~\cite{cogcom} introduce ``Chain of Manipulations'', allowing MLLMs to perform CoT reasoning with external grounding and OCR models, which enable incremental task-solving. Although this approach improves performance, it is limited to zooming in on specific areas and may miss key scattered details. Similarly, Visual CoT~\cite{viscot} enhances performance by focusing on a single region of interest per question, improving both answer accuracy and visual grounding. However, a single grounding and reasoning round is often insufficient for complex, multi-step problems.
To overcome these challenges, we propose two modules: \textbf{\Agent} for reasoning and \textbf{\Grounder} for grounding, enabling multiple rounds of CoT reasoning. This iterative approach allows for better problem-solving by refining both grounding and reasoning across interactions, making it more effective in handling complex tasks, like multi-turn multimodal QAs.

\subsection{Multi-Turn Multimodal Dialogue}

Multi-turn dialogue entails sustained interactions between a human and an MLLM-based agent. These range from casual interactions~\cite{IGC}, to cooperative tasks with shared objectives~\cite{ijcai2024p3} and structured question-answering scenarios~\cite{COCO, TextVQA}. Our focus is on structured question-answering in these dialogues. In language-only multi-turn dialogues, a core challenge lies in managing question interdependence, where responses to earlier questions serve as contextual references in subsequent queries. To provide accurate responses, the model must interpret both the initial answer and the contextual references in follow-up questions. When a visual modality is introduced, the model faces added complexity: it must \ding{182} supplement language information with visual context, \ding{183} synchronize and integrate visual and linguistic data, and \ding{184} manage a reduction in visual focus over prolonged dialogues.

In dialogues where questions are independent, the interdependence challenge is absent, simplifying the interaction to single-turn question answering. Existing multi-turn datasets~\cite{visual_dialog, mmdu, scigraphqa} generally feature QA pairs with minimal interconnection. The MNIST Dialog~\cite{mnist-dialog} dataset incorporates spatial reasoning for correlated QA pairs, but tasks remain relatively simple. ChatterBox~\cite{chatterbox} acknowledges the referential challenge but undermines coherence with rule-based substitutions, simply substituting words occurring repeatedly with ``it'', introducing ambiguities. Our approach addresses these limitations by generating correlated question-answer drafts through rule-based methods, then refining them using GPT-4o-mini~\cite{gpt4omini}. This produces a more complex and realistic multimodal, multi-turn dialogue dataset.
\section{\Dataset: A New Benchmark for Multi-Turn Multimodal Dialogue}
\label{sec:Dataset}

In the following section, we first motivate the choice of scenarios. Next, we show details on how to construct the QA pairs for our \Dataset\ dataset.
We then explain the evaluation process in \Cref{subsec:mmdiag_eval}. 
Finally, we compare \Dataset\ with existing multimodal dialogue datasets in \Cref{subsec:mmdiag_comparison}.
\Dataset\ contains three scenarios: everyday, tabular, and Minigrid. 
Examples of QA pairs are given in \Cref{Appendix:dataset_format}.
Both MMDiag and its generation code will be publicly released.

\subsection{Chosen Scenarios}

The three selected scenarios — \textit{Everyday}, \textit{Tabular}, and \textit{Minigrid} — are chosen to evaluate distinct yet complementary challenges in multimodal reasoning. \textit{Everyday} scenes test common-sense understanding and multi-turn interactions, reflecting real-world AI applications. \textit{Tabular} scenarios require structured data comprehension and numerical reasoning, which many MLLMs struggle with. And \textit{Minigrid} focuses on spatial reasoning and planning, essential for navigation and decision-making. This diverse selection ensures a comprehensive assessment of multimodal understanding. Empirically, all three settings pose significant challenges even for state-of-the-art models like GPT-4o (\Cref{fig:example_minigrid}), with notable failures, such as Visual CoT's inability to generate positive grounding predictions in Tabular tasks (\Cref{tab:gnd}).

\subsection{Dataset Curation}

\noindent\textbf{Everyday Scene Subset.}
The raw source dataset~\cite{Visual_Genome} for this subset contains 108K images, each with detailed annotations.
This allows us to construct a directed graph $\mathcal{G} = (\mathcal{V}, \mathcal{E})$ for each image, where $\mathcal{V}$ represents the objects and $\mathcal{E}$ denotes their relationships.
Then, each QA pair for an image is created and represented as a subgraph of $\mathcal{G}$, \ie, $ \mathcal{G}_{qa} = (\mathcal{V}_{qa}, \mathcal{E}_{qa}) $, with nodes and edges that belong to either the question or the answer.
Note that if a QA pair lacks shared nodes or edges with other subgraphs, we classify it as independent, as it does not contribute to the dialogue’s complexity and does not require information from other QAs for a response.
The created QA pairs are then extended into multi-turn QAs.
We begin by constructing a subgraph pattern $\mathcal{M} = \bigcup_{i=1}^{n} \mathcal{G}_{qa}^i $, where each $ \mathcal{G}_{qa}^i $ represents a subgraph of a QA pair, ensuring $\forall i, \exists j\neq i, \quad s.t. \ \mathcal{V}_{qa}^i \cap \mathcal{V}_{qa}^j \neq \varnothing$. This design guarantees that answering any individual pair requires information from other QA pairs within the multi-turn dialogue. 

We then apply subgraph matching to locate instances of $\mathcal{M}$ in the graph $\mathcal{G}$ for each image, enabling us to create diverse multi-turn QAs. 
We employ GPT-4o-mini~\cite{gpt4omini} to generate various, natural questions, answers, and reasoning steps, while also providing ground truth location data for key objects. 
The specific prompt used in this process is detailed in \Cref{Appendix:everyday_scene}. 

\noindent\textbf{Tabular Scene Subset.} This subset is sourced from ChartQA~\cite{ChartQA}, which contains 18K real-world charts and 23.1K human-authored QA pairs. As ChartQA consists only of single-turn QA, it does not meet our multi-turn dialogue requirements. 
To generate multi-turn question answering, we use GPT-4o-mini, primarily relying on chart images due to the questionable reliability of table-type metadata. To ensure interrelated dialogues, where certain regions are referenced as pronouns to increase complexity, we explicitly emphasize this requirement in the prompt. However, GPT-4o-mini struggles with maintaining this structure, requiring supplementary prompts to guide generation more effectively. Details on the prompt design are provided in \Cref{Appendix:tabular_scene}. 
Finally, we use EasyOCR~\cite{easyocr} to match keywords with corresponding chart regions, enabling generation of bounding boxes for relevant areas.

\noindent\textbf{Minigrid Scene Subset.} Minigrid~\cite{Minigrid} is a Gymnasium-based~\cite{gymnasium} collection of 2D grid-world environments with goal-oriented tasks. The agent, represented as a triangular figure with a discrete action space, navigates maze-like maps and interacts with objects such as doors, keys, and boxes. These tasks test the model’s ability to focus on image details, spatial reasoning, and action planning, with some requiring numerous steps to complete, making them particularly challenging.  
To construct this subset, we use Minigrid and BabyAI~\cite{babyai} to generate grid worlds, tasks, and step-by-step action plans, which are formatted as prompts for GPT-4o-mini. Minigrid creates environments based on specific constraints, saving grid world data as both rendered images and lists of special objects with bounding boxes. BabyAI then identifies feasible solutions by analyzing the agent’s field of view and determining subgoal-aligned actions. To simplify QA generation, we make the entire grid world visible, allowing MLLMs to guide the agent from a top-down perspective. GPT-4o-mini then generates natural questions, reasoning steps, key region queries, and concise final answers. Further details on environment generation and prompt design are in \Cref{Appendix:minigrid}.

\noindent\textbf{Common Visual-Text Subset. }
To enable MLLMs with robust capabilities to answer the question, we also add additional visual-text pairs with high quality from previous works~\cite{llava} to enhance their instruction-following ability.

\subsection{Multi-Turn Multimodal Dialogue Evaluation}
\label{subsec:mmdiag_eval}

The answers in \Dataset\ consist of three main components: the reasoning process, the corresponding grounded key region, and the final answer. 
Accordingly, we evaluate these three components separately.
For the reasoning process and final answer, both of which are expressed in natural language and may vary in phrasing,  we pass the image, question, ground-truth answer, and generated answer to a powerful MLLM for scoring, adhering to widely used evaluation practice. 
To mitigate the potential bias of using the same model for both dataset generation and evaluation, as \Dataset\ is generated using GPT-4o-mini~\cite{gpt4omini}, we instead use Gemini-1.5-Pro~\cite{gemini} in evaluation. Following prior studies~\cite{llm_eval_1, llm_eval_2, llm_eval_3}, we evaluate the MLLMs through ``ad-hoc'' reasoning and scoring across five categories on a 0-10 scale, for greater consistency and interpretability. The complete evaluation prompt is provided in \Cref{Appendix:evaluation}.
Additionally, we use the key region queries and their bounding boxes to constitute a grounding (GND) subset for evaluation.
Since key region queries often involve detailed descriptions of objects or areas, including attributes and relationships, this GND subset can effectively measure grounding capability for complex queries. 
In this context, we use Intersection over Union (IoU) to evaluate the accuracy of grounding.

\subsection{Multimodal Dialogue Datasets Comparison}
\label{subsec:mmdiag_comparison}

We compare \Dataset\ with prior datasets designed for vision-language understanding and reasoning. As shown in \Cref{tab:dataset_comparsion}, \Dataset\ is the first to feature multi-turn, multi-region dialogues with strong QA dependencies, reinforced by a thorough generation process. In contrast, datasets like CB-300k~\cite{chatterbox} and MMDU~\cite{mmdu} lack mechanisms to enforce such dependencies, reducing multi-turn dialogues to mere concatenations of independent QA pairs. Although \Dataset\ has relatively short dialogues, the inherent dependence between turns presents significant challenges for MLLMs, including GPT-4o, as demonstrated in \Cref{fig:example_minigrid}. The grounding and QA test splits include 1,000 unseen images and QA pairs, respectively.

\renewcommand{\arraystretch}{1.2} 
\begin{table*}[ht]
\centering
\setlength{\tabcolsep}{3pt}
\scalebox{0.65}{
\begin{tabular}{l|ccccccc}
\toprule[1pt]
Dataset & QA Scale & GND Scale & Generation Process & Average Turns & Multi-Turn & Multi-Region & Dialogue Correlation \\

\midrule

CB-300k~\cite{chatterbox} & 463k & 254k & GPT-4/Rule-based & 5.49 & \ding{51} & \ding{55} & $\bigcirc$ \\

Visual CoT~\cite{viscot} & 438k & 438k & GPT-4/OCR & 1 & \ding{55} & \ding{55} & \ding{55} \\

CoM~\cite{cogcom} & 76k & - & GPT-4/Tree-Search/Human & 1 & \ding{55}  & $\bigcirc$ & \ding{55} \\

MMDU~\cite{mmdu} & 410k & - & LLM-filtered/GPT-4o & 9 & \ding{51} & \ding{55} & \ding{55} \\

\midrule

\Dataset\ & 639k & 1139k & Graph-search/OCR/GPT-4o-mini & 2.19 & \ding{51}  & \ding{51} & \ding{51} \\

\bottomrule[1pt]
\end{tabular}}
\vspace{-5pt}
\caption{Comparison between \Dataset\ and other multimodal dialogue datasets.
$\bigcirc$: Features are considered, but implemented weakly. }
\addvspace{-5pt}
\label{tab:dataset_comparsion}

\end{table*}

\section{\Model}

In this section, we introduce our proposed \Model\ and its training process.
Using two essential modules named \Agent\ and \Grounder, \Model\ is trained on the train split of \Dataset\ to meet the requirements for multi-turn multimodal dialogue, which provides capabilities of stepwise reasoning and grounding corresponding salient visual regions for each dialogue.

\begin{figure*}[ht]
  \centering
  \includegraphics[width=1\linewidth]{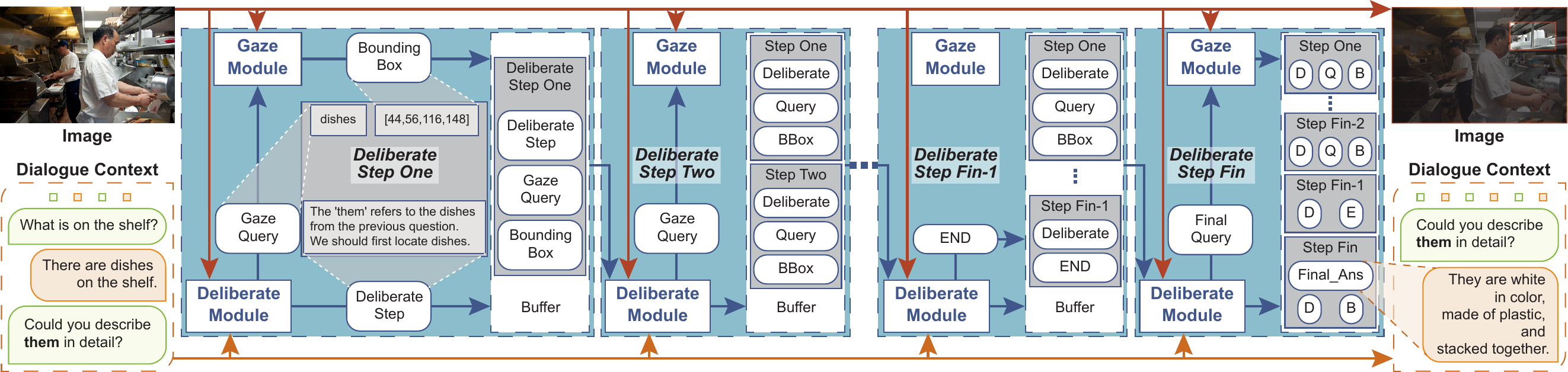}
  \vspace{-15pt}
   \caption{Model architecture of \Model. Regions with blue backgrounds represent a deliberation step and the interaction between the \Agent\ and \Grounder\ modules. At each turn, the \Agent\ module processes the original image, dialogue context, and buffers from both modules. It produces two outputs: (1) a \Agent\ step, stored in the \Agent\ buffer, and (2) a \Grounder\ query, which is processed by the \Grounder\ module. The resulting bounding boxes are then stored in the \Grounder\ buffer.}
   \label{fig:Model}
   \addvspace{-5pt}
\end{figure*}

\subsection{Model Architecture}
\label{subsec:arch}

The overall framework of our model is illustrated in \Cref{fig:Model}.
We use the same architecture (LLaVA-1.5~\cite{llava, llava1.5}) for both \Agent\ and \Grounder\ modules, where the two modules do not share parameters.
Considering the generalization capabilities of MLLMs, we choose not to use a dedicated grounding model like Grounding DINO~\cite{groundingdino} for the \Grounder\ module.
Specifically, the \Agent\ module consists of an LLM as backbone, a pre-trained ViT~\cite{vit} as vision encoder, and one MLP with a projection matrix to serve as the visual-text connector. The same structure applies for the \Grounder\ module, with distinct parameters.
Given an input image $\mathbf{I}_\mathrm{v}$, we consider the entire dialogue contains $T$ turns of questions and answers, which can be represented as $\left(\mathbf{I}_\mathrm{q}^1, \mathbf{I}_\mathrm{a}^1, \cdots, \mathbf{I}_{\mathrm{q}}^T, \mathbf{I}_{\mathrm{a}}^T\right)$, where $\mathbf{I}_\mathrm{q}^t$ and $\mathbf{I}_\mathrm{a}^t$ respectively denote the question and the answer in the $t$-th dialogue turn.

At each turn $t$, given question $\mathbf{I}_{\mathrm{q}}^t$, \Model\ undergoes multiple interactive rounds between the \Agent\ and \Grounder\ modules for reasoning and to generate a reliable response $\mathbf{I}_{\mathrm{a}}^t$. 
To be specific, for the first interactive round, the \Agent\ module $\mathbbm{D}$ takes as input the dialogue context $\mathbf{C}^t=\left(\mathbf{I}_{\mathrm{q}}^1, \mathbf{I}_{\mathrm{a}}^1, \cdots, \mathbf{I}_{\mathrm{q}}^{t-1}, \mathbf{I}_{\mathrm{a}}^{t-1},\mathbf{I}_{\mathrm{q}}^t\right)$ and image $\mathbf{I}_\mathrm{v}$ and outputs a \Agent\ step $\mathbf{S}_{\mathrm{1}}^t$ and a \Grounder\ query $\mathbf{Q}_{\mathrm{1}}^t$. $\mathbf{S}_{\mathrm{1}}^t$ is then stored in the \Agent\ buffer $\mathbf{B}^{t}_d$. The \Grounder\ module $\mathbbm{G}$ takes \Grounder\ query $\mathbf{Q}_{\mathrm{1}}^t$ as input and outputs the corresponding bounding box $\mathbf{o}_{\mathrm{1}}^t$, which is stored in the \Grounder\ buffer $\mathbf{B}^t_g$. 
In each subsequent interactive round $i$ of \Agent\ and \Grounder, the \Agent\ module takes as input the image $\mathbf{I}_\mathrm{v}$, the dialogue context  $\mathbf{C}^t$, the \Grounder\ buffer $\mathbf{B}^t_g=\left(\mathbf{o}_{1}^t, \cdots, \mathbf{o}_{i-1}^t\right)$, and the \Agent\ buffer $\mathbf{B}^t_d=\left(\mathbf{S}_{1}^t, \cdots, \mathbf{S}_{i-1}^t\right)$ to generate \Agent\ step $\mathbf{S}_{i}^t$ and \Grounder\ query $\mathbf{Q}_{i}^t$. 
The \Grounder\ module $\mathbbm{G}$, again, takes \Grounder\ query $\mathbf{Q}_{i}^t$ as input and outputs the annotation bounding box $\mathbf{o}_{i}^t$. This process continues until the \Agent\ module outputs `END' as the \Grounder\ query $\mathbf{Q}_{\mathrm{Fin}-1}^k$, indicating that the \Agent\ and \Grounder\ back-and-forth process is complete.

Finally, the image, the dialogue context, and all the buffers are fed into the \Agent\ module $\mathbbm{D}$ to produce the final answer $\mathbf{S}_{\mathrm{Fin}}^t$ (i.e., $\mathbf{I}_{\mathrm{a}}^t$) and the \Grounder\ query $\mathbf{Q}_{\mathrm{Fin}}^t$. The \Grounder\ module $\mathbbm{G}$ then provides the bounding box of the salient area $\mathbf{o}_{\mathrm{Fin}}^t$ for the $t$-th dialogue turn. The final output is $\mathbf{S}_{\mathrm{Fin}}^t$, along with the optional key region bounding box $\mathbf{o}_{\mathrm{Fin}}^t$, as well as the \Agent\ process $\left(\mathbf{S}_{\mathrm{1}}^t, \cdots, \mathbf{S}_{\mathrm{Fin}-1}^t\right)$, if required. The final answer $\mathbf{I}_{\mathrm{a}}^t$ is then appended to the dialogue context for the next dialogue turn.

\subsection{Model Training}
\label{subsec:training}

\renewcommand{\arraystretch}{1.2} 
\begin{table*}[ht]
\centering
\setlength{\tabcolsep}{3pt}
\scalebox{0.85}{
\begin{tabular}{l|l|ccc|cc|c}
\toprule
\multirow{3}{*}{ \textbf{Model}} &
\multirow{3}{*}{ \textbf{Train Data}} &
\multicolumn{3}{c}{ \textbf{\Dataset\ GND Testset}} &
\multicolumn{2}{|c|}{\textbf{GND Dataset}} & \multicolumn{1}{|c}{\multirow{2}{*}{ \textbf{Average}}} \\ 
  \cmidrule(lr){3-5}
  \cmidrule(lr){6-7}
    &
    &
  { \textbf{Everyday}} &
  { \textbf{Tabular}} &
  { \textbf{Minigrid}} &
   MSCOCO &
   RefCOCO \\
\midrule
Grounding DINO~\cite{groundingdino} & - & 0.384 & 0.001 & 0.209 & \textbf{0.715} & 0.469 & 0.356\ \\ 
LLaVA~\cite{llava} & LCS558K+Mixed665K &
  0.237 & 0.006 & 0.142 & 0.365 & 0.414 & 0.233\ \\ 
 Visual CoT~\cite{viscot}   &
 VisCoT  & 0.220  &  0.003 &  0.160 & 0.321 & 0.362 & 0.213  \ \\ 
\midrule
\Model\ & COCO  & 0.307 &  0.008 & 0.199 & 0.662 & 0.765 & 0.388 \\ 
\Model\ & MMDiag  & 0.369 &  0.466 & \textbf{1.0} & 0.259 & 0.257 & 0.471\\ 
\Model\ & MMDiag + COCO  & 0.399 & \textbf{0.487} & 0.988 & 0.624 & 0.742 & 0.648 \\ 
\Model\ & MMDiag + COCO + VisCoT & \textbf{0.433} & 0.281 & 0.910& \textbf{0.662} & \textbf{0.837} & 0.625 \\
\bottomrule
\end{tabular}
}
\vspace{-5pt}
\caption{
Comparison results with existing MLLMs on Grounding benchmarks (GND) to 
demonstrate the challenging characteristics of our dataset \Dataset. 
We use Intersection over Union (IoU) as the evaluation metric.}
\label{tab:gnd}
\addvspace{-10pt}
\end{table*}

The training process of both \Agent\ and \Grounder\ modules follows that of LLaVA, and \Model\ provides two prompt templates $p^{\mathrm{d}}$ and $p^{\mathrm{g}}$ for \Agent\ and \Grounder\ respectively. At the $i$-th round of \Agent\ and \Grounder\ for Question $\mathbf{I}_{\mathrm{q}}^{t}$, the instruction $\mathbf{R}\mathrm{in}_i^{\mathrm{d}}$ for the \Agent\ module is: 
\vspace{-7pt}
\begin{equation}
    \mathbf{R}\mathrm{in}_i^{\mathrm{d}}= \begin{cases} p^{\mathrm{d}}{\left(\mathbf{I}_\mathrm{v}, \mathbf{C}^t\right)}, \quad i=1 
    \\ p^{\mathrm{d}}{\left(\mathbf{I}_\mathrm{v}, \mathbf{C}^t, \mathbf{B}^t_g, \mathbf{B}^t_d\right)}, \quad 1<i<\mathrm{Fin}
    \\ p^{\mathrm{d}}{\left(\mathbf{I}_\mathrm{v}, \mathbf{C}^t, \mathbf{B}^t_g, \mathbf{B}^t_d, \mathrm{Fin}\right)}, \quad i=\mathrm{Fin},
    \end{cases}
    \label{eq:Agent}
\end{equation}
where $\mathbf{B}^t_d=\left(\mathbf{S}_{{1}}^t, \cdots, \mathbf{S}_{{i-1}}^{t}\right)$ and $\mathbf{B}^t_g=\left(\mathbf{Q}_{{1}}^t, \cdots, \mathbf{Q}_{{i-1}}^{t}\right)$. The instruction $\mathbf{R}\mathrm{in}_{i}^{\mathrm{g}}$ for the \Grounder\ module is:
\begin{equation}
    \mathbf{R}\mathrm{in}_{{i}}^{\mathrm{g}}= p^{\mathrm{g}}{\left(\mathbf{I}_\mathrm{v}, \mathbf{Q}_{{i}}^t\right)}, \quad i\leq\mathrm{Fin}, i\neq{\mathrm{Fin}-1}.
    \label{eq:gaze}
\end{equation}
We fine-tune the LLM on the prediction tokens, utilizing the auto-regressive training objective to optimize. We compute the probability of the target output $\mathbf{R}\mathrm{out}_i^\mathrm{x}$ with length $L$ at $i$-th round by:

\vspace{-10pt}
\begin{equation}\label{eq:output}
\begin{split}
    p\left(\mathbf{R}\mathrm{out}_{{i}}^{\mathrm{x}} \mid \mathbf{R}\mathrm{in}_{{i}}^{\mathrm{x}}\right)&=
    \prod_{l=1}^L p_{\boldsymbol{\theta}^\mathrm{x}}\left(r_l \mid \mathbf{R}{\mathrm{in}}_i^{\mathrm{x}}, \mathbf{R}{\mathrm{out}}_{,<l}^{\mathrm{x}}\right), \\
    &\quad\quad\quad\quad\text { where }\mathrm{x} \in \{ \mathrm{d}, \mathrm{g}\}.
\end{split}
\end{equation}

$\boldsymbol{\theta}^\mathrm{x}$ is the trainable parameters of the \Agent\ and \Grounder\ modules respectively, with $\mathrm{x} \in \{ \mathrm{d}, \mathrm{g} \}$. $\mathbf{R}{\mathrm{in}}_{i}^{\mathrm{x}}$ are the input tokens of $i$-th round of the \Agent\ and \Grounder\ interaction process. $\mathbf{R}{\mathrm{out}}_{,<l}^{\mathrm{x}}$ are the answer tokens before the current prediction token $r_l$.

Our \Agent\ and \Grounder\ modules take LLaVA-1.5~\cite{llava1.5} as base model. For the \Grounder\ module, since grounding such salient areas as words and objects with detailed descriptions is quite challenging, we can first fine-tune it with an additional grounding dataset, and then fine-tune \Agent\ and \Grounder\ modules together. We combine the fine-tuning dataset from LLaVA~\cite{llava} and the grounding datasets of MSCOCO~\cite{COCO} and RefCOCO~\cite{refcoco,chatterbox} with the augmentation grounding split of \Dataset\ to generate the grounding dataset; and we also combine the fine-tuning dataset from LLaVA with the training split of the \Dataset\ dataset to generate the entire training dataset. For data points in LLaVA, \Model\ does not add \Agent\ prompts for the \Agent\ module, thus instructing the \Agent\ module to maintain the ability to output answers in general format.
\section{Experiments}

\subsection{Implementation Details}

We use LLaVA-1.5-7B~\cite{llava1.5} as the foundation model for both \Agent\ and \Grounder\ modules, with CLIP-ViT-Large-Patch14-336~\cite{vit} as vision tower.  
Training is conducted on 8 × A800 GPUs with a learning rate of 2e-5. \Agent\ and \Grounder\ are optimized separately via supervised learning with ground-truth outputs per round. During inference, the \Grounder\ module signals reasoning completion by outputting ``END'' for turn \(T_x\) (\Cref{tab:chartqa_round}), with the round number dynamically determined by \Model. Additional training details are provided in the \Cref{Appendix:model,Appendix:implementation}.

\subsection{Results on \Dataset\ }

\begin{table*}[ht]
\centering
\scalebox{0.8}{
\begin{tabular}{l|c|l|cccccc|c}
\toprule
\multirow{3}{*}{ \textbf{Model}} &  \multirow{3}{*}{ \textbf{\Grounder\ }} & \multirow{3}{*}{ \textbf{Train Data}} &  \multicolumn{6}{c|}{ \textbf{\Dataset\ }} & \multirow{3}{*}{ \textbf{Average}}
\\ 
\cmidrule{4-9} & & & \multicolumn{2}{c}{ \textbf{Everyday}} & \multicolumn{2}{c}{ \textbf{Tabular}} &  \multicolumn{2}{c|}{ \textbf{Minigrid}}
\\ 
\cmidrule(lr){4-5} \cmidrule(lr){6-7} \cmidrule(lr){8-9} 
& & & reasoning & answer & reasoning & answer & reasoning & answer \\
\midrule

LLaVA~\cite{llava} &  \ding{55} & LCS558K+Mixed665K & 2.55	& 4.85&1.00&1.28&2.29&0.42&2.21 \\ 
CogCoM~\cite{cogcom} & \ding{55} & - & 3.05	& 5.45&	0.50&	1.25&	0.53&	0.96
&2.20 \\ 
Visual CoT~\cite{viscot} & \ding{55} & VisCoT & 4.15&4.90&1.23&1.95&1.09&2.50&2.81 \\ 
\midrule
\Model &  \ding{55} & MMDiag &  4.25&4.95&3.61&4.20&4.95&4.27&4.32  \\ 
\Model & \ding{51} & MMDiag & 5.82&\textbf{6.15}&3.95&4.05&5.10&4.15&4.92 \\ 
\Model & \ding{51} & MMDiag+COCO & 6.35&5.97&3.95&4.30&5.75&4.93&5.18 \\ 
\Model & \ding{51} & GT & \textbf{6.85}&5.80&\textbf{6.32}&\textbf{7.76}&\textbf{7.37}&\textbf{9.15}&\textbf{7.00} \\ 
\bottomrule[1pt]
\end{tabular}}
\vspace{-5pt}
\caption{Comparison of the evaluation score with baselines to validate the \Grounder\ module, we use Gemini-1.5-Pro to evaluate the performance of the reasoning process and the final answer. The evaluation process is detailed in~\Cref{subsec:mmdiag_eval}.
}
\addvspace{-5pt}
\label{tab:ans}
\end{table*}

\begin{table}[ht]
\centering
\scalebox{0.8}{
\begin{tabular}{l|cccccccc}
\toprule
\multirow{3}{*}{\textbf{Model}} &
\multicolumn{8}{c}{\textbf{Tabular}} \\ 
\cmidrule{2-9}
&
\multicolumn{4}{c}{\textbf{Reasoning}} &
\multicolumn{4}{c}{\textbf{Answer}} \\ 
\cmidrule(lr){2-5} \cmidrule(lr){6-9} 
&
T1 & T2 & T3 & T4 & T1 & T2 & T3 & T4 \\ 
\midrule
CogCoM  & 0.55 & 0.91 & 1.15 & 0.67 & 1.75 & 0.73 & 0.85 & 0.35 \\ 
Visual CoT  & 1.50 & 1.05 & 1.33 & 1.02 & 1.86 & 1.24 & 1.03 & 0.88 \\ 
LLaVA  & 2.34 & 0.35 & 1.00 & 0.58 & 1.42 & 0.50 & 0.97 & 0.50 \\ 
\midrule
w/o \Grounder\ & \textbf{4.01} & 3.05 & 2.15 & 1.66 & \textbf{3.47} & 2.03 & 1.65 & 1.63 \\ 
with \Grounder\ & 3.86 & \textbf{3.34} & \textbf{2.31} & \textbf{2.53} & 3.25 & \textbf{2.65} & \textbf{2.17} & \textbf{1.98} \\ 
\bottomrule
\end{tabular}
}
\vspace{-5pt}
\caption{The Gemimi-1.5-Pro evaluation of the reasoning process and the final answer, scaling to 0-10, at turns 1 to 4 under the tabular scenario, where T$*$ denotes the $*$-th turn in the dialogue.}
\label{tab:chartqa_round}
\addvspace{-10pt}
\end{table}

\subsubsection{Visual Grounding}
In this section, we focus primarily on how the \Dataset\ dataset benefits the grounding performance of MLLMs.
Grounding is a crucial capability for MLLMs, enabling them to focus on relevant salient regions and reveal the visible reasoning process in dialogues, rather than functioning as a black box.
We evaluate our \Model\ on several general grounding (GND) benchmarks~\cite{COCO,refcoco,chatterbox}, as well as our \Dataset\ GND benchmark. 
We use the average Intersection over Union (IoU) scores as the metric to assess GND performance, 
with results summarized in \Cref{tab:gnd}. 
When comparing \Model's performance on established GND benchmarks like MSCOCO to its performance on \Dataset, we observe a significant decline on \Dataset, highlighting its increased difficulty relative to existing benchmarks.
Existing models, such as Visual CoT, incorporate regions of interest for multimodal dialogue learning, but perform unsatisfactorily on GND datasets.
For instance, Visual CoT scores $\textbf{-0.394}$ compared to Grounding DINO on the MSCOCO benchmark and performs worse than LLaVA.
These results indicate a lack of robustness in explicitly grounding relevant areas in images.
In contrast, \Model, trained with limited standard GND annotations provided by \Dataset\ and MSCOCO, shows significant improvements in both MSCOCO and RefCOCO benchmarks and performs better across the three subset scenarios of \Dataset. 
Notably, the MSCOCO data here is used solely to enhance grounding capability, and we intentionally limit the scale of GND data to prevent dataset size from influencing our conclusion. 
As shown in Row 4, \Model\ performs the poorest when trained exclusively on the MSCOCO dataset, underscoring the necessity and benefits of our \Dataset\ dataset.

\subsubsection{Multi-Turn Reasoning}

We also evaluate our model's multi-turn reasoning capabilities using the \Dataset\ benchmark. 
Beyond evaluating the correctness of the final answers, the evaluator also assesses the coherence and logic of the reasoning process within the \Agent\ module. Detailed results are presented in \Cref{tab:ans}. 
``GT'' denotes scenarios where the \Agent\ module receives ground-truth inputs for the reasoning step, serving as an upper bound. Except for the GT results, \Grounder\ queries are generated by \Model, preventing information leakage. As expected, the GT setting significantly outperforms other settings, highlighting considerable room for improvement. 
To validate the effectiveness of our proposed module, we observe that the \Grounder\ module enhances performance in specific reasoning tasks. 
For instance, in the everyday scenario, models utilizing the \Grounder\ module achieve notably higher accuracy than those without it, demonstrating its ability to enhance focus and accuracy in reasoning. 
When there are multiple things, of the same kind, with different locations and attributes, in the image, chances are that the model cannot tell which object is exactly the one mentioned in the question. 
If the specific target is annotated on the image, the model can regain focus on it easily and avoid such cases that it fails to locate the right target when the reasoning process moves on.

To further evaluate model performance, we compare our \Model\ with CogCoM~\cite{cogcom} and Visual CoT~\cite{viscot}, both of which can focus on specific regions and manage multimodal dialogues.
Results show that \Model\ has significant advantages, especially in the tabular and Minigrid scenarios, reflecting the complexity of the \Dataset\ dataset and the strengths of \Model's architecture, featured with the \Agent\ and \Grounder\ modules. 
To deepen the analysis, we show a comparison of results in tabular scenes under different numbers of dialogue turns in \Cref{tab:chartqa_round}.
\Model\ consistently outperforms the other models in the second, third, and fourth rounds under the tabular scenario, underscoring its superior capability in handling long-context scenes with contextual and pronoun references.
Meanwhile, the \Grounder\ module shows more significant improvement, especially for increasingly long dialogues (\eg T3 or T4), which further validates its effectiveness and benefits in long-context multimodal understanding. 
It is important to note that \Dataset's tabular scenes in \Cref{tab:ans} include QA pairs of varying lengths (2–4), while \Cref{tab:chartqa_round} focuses only on dialogues with exactly 4 QA pairs.

\begin{figure*}[ht]
  \centering
  \includegraphics[width=0.95\linewidth]{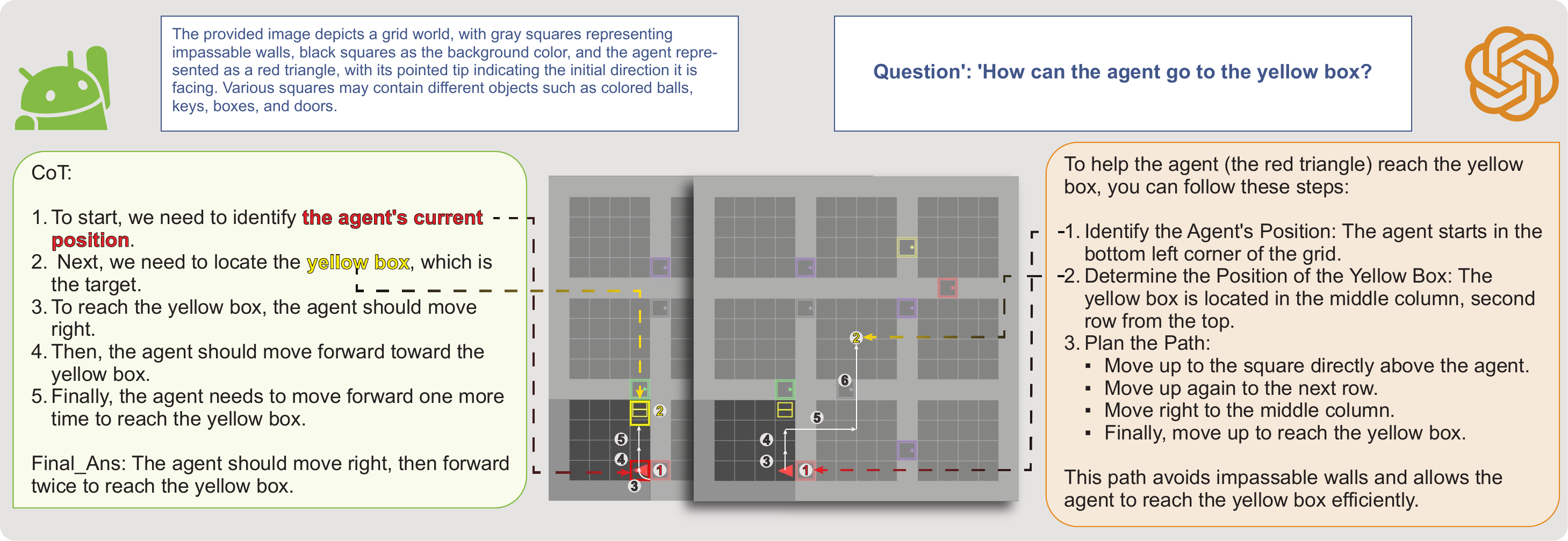}
   \caption{Comparison for an example of the Minigrid scenario, one of the subsets in \Dataset. 
   We give \Model\ (green) and GPT-4o (orange) the same environmental description and question. \Model\ focuses on the key regions and gives the correct reasoning process and the final answer. In contrast, GPT-4o fails to locate the object and thus gives the wrong answer. 
   Examples for the \Dataset\ subsets of everyday scenarios and tabular scenes can be found in \Cref{Appendix:qual_cereb}.}
   \label{fig:example_minigrid}
   \addvspace{-5pt}
\end{figure*}

\subsection{Qualitative Results. }
In this section, we provide additional examples of the visual grounding and reasoning capabilities of \Model. More visualization results can be found in \Cref{Appendix:qual_gaze,Appendix:qual_cereb}.

\begin{figure}
	\centering
	\begin{subfigure}{0.48\linewidth}
		\centering
		\includegraphics[width=1\linewidth]{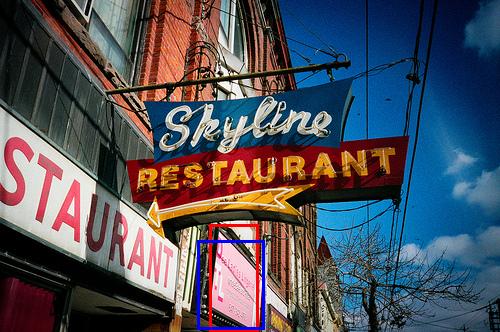}
		\caption{\Model}
		\label{subfig:Model_gazing}
	\end{subfigure}
        \begin{subfigure}{0.48\linewidth}
		\centering
		\includegraphics[width=1\linewidth]{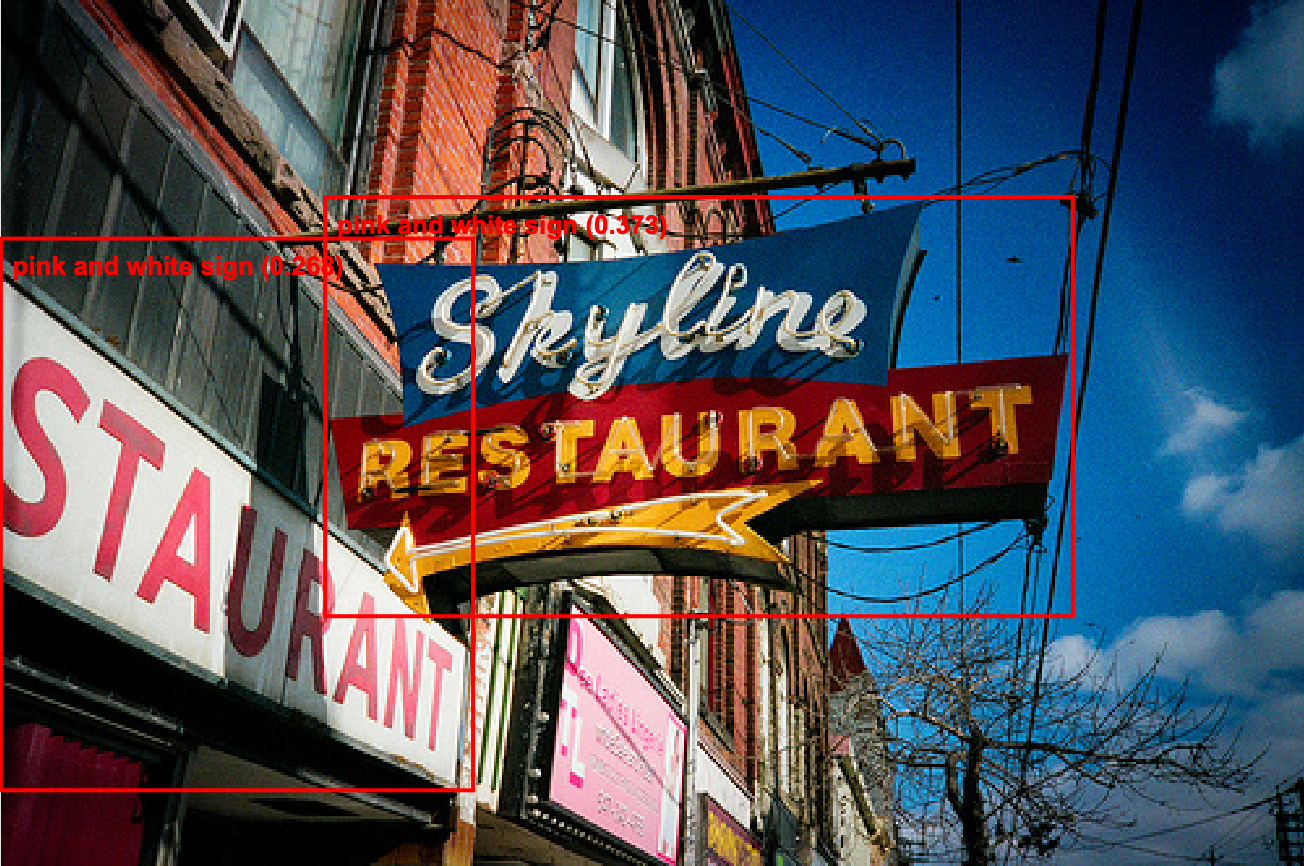}
		\caption{Grounding DINO}
		\label{subfig:DINO_gazing}
	\end{subfigure}
    \vspace{-5pt}
	\caption{A grounding comparison between Grounding DINO and \Model's \Grounder\ module
, with the \Grounder\ query ``pink and white sign''. In (a), the red bounding box represents the ground-truth answer, while the blue one indicates the output generated by the \Grounder\ module in \Model. In (b), the red bounding boxes show the outputs produced by Grounding DINO. }
	\label{fig:gazing_vg}
    \addvspace{-5pt}
\end{figure}

\noindent\textbf{Visual Grounding.}
The \Grounder\ module offers both grounding and OCR capabilities across diverse scenarios. 
As illustrated in \Cref{subfig:DINO_gazing}, Grounding DINO~\cite{groundingdino} struggles in complex scenes where multiple objects of the same category exist with different attributes or relationships, therefore often failing to locate the target object precisely. 
In contrast, \Model's \Grounder\ module effectively manages such situations, as shown in \Cref{subfig:Model_gazing}.
Additionally, when faced with tasks requiring text recognition, the \Grounder\ module exhibits more robust OCR capabilities, accurately identifying and localizing specific keywords.

\noindent \textbf{Multi-Turn Reasoning. }
With the incorporation of the \Grounder\ module, our model can also more effectively focus on fine-grained details distributed across the image, offering a clear advantage in tasks that demand cohesive reasoning across both visual and linguistic information.
As shown in \Cref{fig:example_minigrid}, a comparison between our \Model\ and GPT-4o within a simple Minigrid environment highlights this benefit. 
Despite detailed descriptions provided in the prompt, GPT-4o struggles with completing a short-range, single-subgoal task, underscoring the strengths of our dataset and methodology.

\subsection{Ablation Study}

We observe a counterintuitive performance trend when comparing \Model\ with and without the \Grounder\ module. To analyze its impact, we fine-tune \Model\ and Visual CoT on \Dataset\ and confirm \Grounder’s effectiveness. However, its gains are limited, likely due to low-resolution image inputs.  
Failure cases show that when dialogues reference tiny key regions (under 0.2\% of the image), \Grounder\ often produces inaccurate bounding boxes, confusing the \Agent\ module. The CLIP-ViT-Large-Patch14-336 encoder further limits resolution, contributing to errors.  
On standard multimodal benchmarks, \Model\ performs comparably or slightly lower, as it targets complex multi-region dialogues without in-domain training data. Ablation details are in \Cref{Appendix:ablation}. 
\section{Conclusion}
In this paper, we focus on a key challenging task scenario for MLLMs\textemdash multi-turn multimodal dialogue.
To address it, we first introduce a specially designed dataset, \Dataset, where accomplishing tasks requires properly integrating visual information across different regions of an image and connecting multimodal information across various QA pairs. 
This setting closely resembles natural conversations and poses significant challenges to current MLLMs.
To solve this, we construct the \Dataset\ dataset across three distinct scenarios—everyday, tabular, and Minigrid—using a combination of rule-based methods and GPT-4o-mini to ensure robustness and diversity.
Experimental results highlight the challenges posed by \Dataset.
Therefore, we propose \Model, an MLLM inspired by human visual processing, composed of two primary modules: \Grounder\ and \Agent. 
The \Agent\ module performs CoT reasoning step by step, with the assistance of the \Grounder\ module, which provides annotations of salient regions to focus on.
Experiments show that this design enhances both grounding and reasoning capabilities, effectively addressing \Dataset\ challenges.
We hope our work contributes to advancing the development of more intelligent MLLMs.

\vspace{-2mm}
\section*{Limitations}
Although \Dataset\ contains diverse data, our methods can be expected to generate even more scenarios and complex questions, resulting in even more challenging datasets for multi-turn multimodal dialogue.
While qualitative results and case studies demonstrate the effectiveness of our approach, there remains considerable room for improvement. 
The potential performance drops with the introduction of \Grounder\ module may stem from failures in queries involving extremely tiny objects. Fine-tuning \Grounder\ to abstain from answering when uncertain or replacing the vision encoder backbone may enhance its robustness. Further exploration of training paradigms and model architecture could also potentially lead to enhanced performance. 

{
    \small
    \bibliographystyle{ieeenat_fullname}
    \bibliography{arxiv}

\begin{thebibliography}{51}
\providecommand{\natexlab}[1]{#1}
\providecommand{\url}[1]{\texttt{#1}}
\expandafter\ifx\csname urlstyle\endcsname\relax
  \providecommand{\doi}[1]{doi: #1}\else
  \providecommand{\doi}{doi: \begingroup \urlstyle{rm}\Url}\fi

\bibitem[Achiam et~al.(2023)Achiam, Adler, Agarwal, Ahmad, Akkaya, Aleman,
  Almeida, Altenschmidt, Altman, Anadkat, et~al.]{gpt4report}
Josh Achiam, Steven Adler, Sandhini Agarwal, Lama Ahmad, Ilge Akkaya,
  Florencia~Leoni Aleman, Diogo Almeida, Janko Altenschmidt, Sam Altman,
  Shyamal Anadkat, et~al.
\newblock {Gpt-4 technical report}.
\newblock \emph{arXiv preprint arXiv:2303.08774}, 2023.

\bibitem[Bai et~al.(2023{\natexlab{a}})Bai, Bai, Chu, Cui, Dang, Deng, Fan, Ge,
  Han, Huang, Hui, Ji, Li, Lin, Lin, Liu, Liu, Lu, Lu, Ma, Men, Ren, Ren, Tan,
  Tan, Tu, Wang, Wang, Wang, Wu, Xu, Xu, Yang, Yang, Yang, Yang, Yao, Yu, Yuan,
  Yuan, Zhang, Zhang, Zhang, Zhang, Zhou, Zhou, Zhou, and Zhu]{qwen}
Jinze Bai, Shuai Bai, Yunfei Chu, Zeyu Cui, Kai Dang, Xiaodong Deng, Yang Fan,
  Wenbin Ge, Yu Han, Fei Huang, Binyuan Hui, Luo Ji, Mei Li, Junyang Lin, Runji
  Lin, Dayiheng Liu, Gao Liu, Chengqiang Lu, Keming Lu, Jianxin Ma, Rui Men,
  Xingzhang Ren, Xuancheng Ren, Chuanqi Tan, Sinan Tan, Jianhong Tu, Peng Wang,
  Shijie Wang, Wei Wang, Shengguang Wu, Benfeng Xu, Jin Xu, An Yang, Hao Yang,
  Jian Yang, Shusheng Yang, Yang Yao, Bowen Yu, Hongyi Yuan, Zheng Yuan,
  Jianwei Zhang, Xingxuan Zhang, Yichang Zhang, Zhenru Zhang, Chang Zhou,
  Jingren Zhou, Xiaohuan Zhou, and Tianhang Zhu.
\newblock {Qwen Technical Report}.
\newblock \emph{arXiv preprint arXiv:2309.16609}, 2023{\natexlab{a}}.

\bibitem[Bai et~al.(2023{\natexlab{b}})Bai, Bai, Yang, Wang, Tan, Wang, Lin,
  Zhou, and Zhou]{qwenvl}
Jinze Bai, Shuai Bai, Shusheng Yang, Shijie Wang, Sinan Tan, Peng Wang, Junyang
  Lin, Chang Zhou, and Jingren Zhou.
\newblock {Qwen-VL: A Versatile Vision-Language Model for Understanding,
  Localization, Text Reading, and Beyond}.
\newblock \emph{arXiv preprint arXiv:2308.12966}, 2023{\natexlab{b}}.

\bibitem[Brown(2020)]{ICL}
Tom~B. Brown.
\newblock {Language Models are Few-Shot Learners}.
\newblock \emph{arXiv preprint arXiv:2005.14165}, 2020.

\bibitem[Changpinyo et~al.(2021)Changpinyo, Sharma, Ding, and Soricut]{cc12m}
Soravit Changpinyo, Piyush Sharma, Nan Ding, and Radu Soricut.
\newblock {Conceptual 12M: Pushing Web-Scale Image-Text Pre-Training To
  Recognize Long-Tail Visual Concepts}.
\newblock In \emph{CVPR}, pages 3558--3568, 2021.

\bibitem[Chen et~al.(2024{\natexlab{a}})Chen, Dong, Shu, Zhang, Sesay,
  Karlsson, Fu, and Shi]{ijcai2024p3}
Guangyao Chen, Siwei Dong, Yu Shu, Ge Zhang, Jaward Sesay, Börje~F. Karlsson,
  Jie Fu, and Yemin Shi.
\newblock {AutoAgents: A Framework for Automatic Agent Generation}.
\newblock In \emph{IJCAI}, 2024{\natexlab{a}}.

\bibitem[Chen et~al.(2024{\natexlab{b}})Chen, Chen, Liu, Jiang, and
  Wang]{llm_eval_3}
Guiming~Hardy Chen, Shunian Chen, Ziche Liu, Feng Jiang, and Benyou Wang.
\newblock {Humans or llms as the judge? a study on judgement biases}.
\newblock \emph{arXiv preprint arXiv:2402.10669}, 2024{\natexlab{b}}.

\bibitem[Chen et~al.(2023)Chen, Djolonga, Padlewski, Mustafa, Changpinyo, Wu,
  Ruiz, Goodman, Wang, Tay, et~al.]{palix}
Xi Chen, Josip Djolonga, Piotr Padlewski, Basil Mustafa, Soravit Changpinyo,
  Jialin Wu, Carlos~Riquelme Ruiz, Sebastian Goodman, Xiao Wang, Yi Tay, et~al.
\newblock {PaLI-X: On Scaling up a Multilingual Vision and Language Model}.
\newblock \emph{arXiv preprint arXiv:2305.18565}, 2023.

\bibitem[Chevalier-Boisvert et~al.(2019)Chevalier-Boisvert, Bahdanau, Lahlou,
  Willems, Saharia, Nguyen, and Bengio]{babyai}
Maxime Chevalier-Boisvert, Dzmitry Bahdanau, Salem Lahlou, Lucas Willems,
  Chitwan Saharia, Thien~Huu Nguyen, and Yoshua Bengio.
\newblock {BabyAI: First Steps Towards Grounded Language Learning With a Human
  In the Loop}.
\newblock In \emph{ICLR}, 2019.

\bibitem[Chevalier-Boisvert et~al.(2023)Chevalier-Boisvert, Dai, Towers,
  de~Lazcano, Willems, Lahlou, Pal, Castro, and Terry]{Minigrid}
Maxime Chevalier-Boisvert, Bolun Dai, Mark Towers, Rodrigo de Lazcano, Lucas
  Willems, Salem Lahlou, Suman Pal, Pablo~Samuel Castro, and Jordan Terry.
\newblock {Minigrid \& Miniworld: Modular \& Customizable Reinforcement
  Learning Environments for Goal-Oriented Tasks}.
\newblock \emph{CoRR}, abs/2306.13831, 2023.

\bibitem[Chiang et~al.(2023)Chiang, Li, Lin, Sheng, Wu, Zhang, Zheng, Zhuang,
  Zhuang, Gonzalez, Stoica, and Xing]{vicuna}
Wei-Lin Chiang, Zhuohan Li, Zi Lin, Ying Sheng, Zhanghao Wu, Hao Zhang, Lianmin
  Zheng, Siyuan Zhuang, Yonghao Zhuang, Joseph~E. Gonzalez, Ion Stoica, and
  Eric~P. Xing.
\newblock {Vicuna: An Open-Source Chatbot Impressing GPT-4 with 90\%* ChatGPT
  Quality}, 2023.

\bibitem[Cursor(2024)]{cursor}
Cursor.
\newblock {The AI Code Editor}.
\newblock \url{https://www.cursor.com/}, 2024.

\bibitem[Das et~al.(2017)Das, Kottur, Gupta, Singh, Yadav, Moura, Parikh, and
  Batra]{visual_dialog}
Abhishek Das, Satwik Kottur, Khushi Gupta, Avi Singh, Deshraj Yadav, Jose M.~F.
  Moura, Devi Parikh, and Dhruv Batra.
\newblock {Visual Dialog}.
\newblock In \emph{CVPR}, 2017.

\bibitem[DeepL(2024)]{deepl}
DeepL.
\newblock {Better writing with DeepL Write}.
\newblock \url{https://www.deepl.com/en/write}, 2024.

\bibitem[Driess et~al.(2023)Driess, Xia, Sajjadi, Lynch, Chowdhery, Ichter,
  Wahid, Tompson, Vuong, Yu, et~al.]{palme}
Danny Driess, Fei Xia, Mehdi~SM Sajjadi, Corey Lynch, Aakanksha Chowdhery,
  Brian Ichter, Ayzaan Wahid, Jonathan Tompson, Quan Vuong, Tianhe Yu, et~al.
\newblock {PaLM-E: An Embodied Multimodal Language Model}.
\newblock \emph{arXiv preprint arXiv:2303.03378}, 2023.

\bibitem[Feng et~al.(2024)Feng, Wang, Liu, Zheng, and Lu]{feng2024llama}
Yicheng Feng, Yuxuan Wang, Jiazheng Liu, Sipeng Zheng, and Zongqing Lu.
\newblock {LLaMA-Rider: Spurring Large Language Models to Explore the Open
  World}.
\newblock In \emph{NAACL}, pages 4705--4724, 2024.

\bibitem[JaidedAI(2024)]{easyocr}
JaidedAI.
\newblock {EasyOCR}.
\newblock \url{https://github.com/JaidedAI/EasyOCR}, 2024.

\bibitem[Kazemzadeh et~al.(2014)Kazemzadeh, Ordonez, Matten, and Berg]{refcoco}
Sahar Kazemzadeh, Vicente Ordonez, Mark Matten, and Tamara Berg.
\newblock {ReferItGame: Referring to Objects in Photographs of Natural Scenes}.
\newblock In \emph{EMNLP}, pages 787--798, 2014.

\bibitem[Krishna et~al.(2017)Krishna, Zhu, Groth, Johnson, Hata, Kravitz, Chen,
  Kalantidis, Li, Shamma, et~al.]{Visual_Genome}
Ranjay Krishna, Yuke Zhu, Oliver Groth, Justin Johnson, Kenji Hata, Joshua
  Kravitz, Stephanie Chen, Yannis Kalantidis, Li-Jia Li, David~A Shamma, et~al.
\newblock {Visual Genome: Connecting Language and Vision Using Crowdsourced
  Dense Image Annotations}.
\newblock \emph{IJCV}, 123:\penalty0 32--73, 2017.

\bibitem[Lee et~al.(2024)Lee, Hong, and Thorne]{llm_eval_1}
Noah Lee, Jiwoo Hong, and James Thorne.
\newblock {Evaluating the Consistency of LLM Evaluators}.
\newblock \emph{arXiv preprint arXiv:2412.00543}, 2024.

\bibitem[Li et~al.(2023)Li, Li, Savarese, and Hoi]{blip2}
Junnan Li, Dongxu Li, Silvio Savarese, and Steven Hoi.
\newblock {BLIP-2: Bootstrapping Language-Image Pre-training with Frozen Image
  Encoders and Large Language Models}.
\newblock In \emph{ICML}, pages 19730--19742. PMLR, 2023.

\bibitem[Li and Tajbakhsh(2023)]{scigraphqa}
Shengzhi Li and Nima Tajbakhsh.
\newblock {SciGraphQA: A Large-Scale Synthetic Multi-Turn Question-Answering
  Dataset for Scientific Graphs}.
\newblock \emph{arXiv preprint arXiv:2308.03349}, 2023.

\bibitem[Lin et~al.(2014)Lin, Maire, Belongie, Hays, Perona, Ramanan,
  Doll{\'a}r, and Zitnick]{COCO}
Tsung-Yi Lin, Michael Maire, Serge Belongie, James Hays, Pietro Perona, Deva
  Ramanan, Piotr Doll{\'a}r, and C~Lawrence Zitnick.
\newblock {Microsoft COCO: Common Objects in Context}.
\newblock In \emph{ECCV}, pages 740--755. Springer, 2014.

\bibitem[Liu et~al.(2024{\natexlab{a}})Liu, Li, Li, and Lee]{llava1.5}
Haotian Liu, Chunyuan Li, Yuheng Li, and Yong~Jae Lee.
\newblock {Improved Baselines with Visual Instruction Tuning}.
\newblock In \emph{CVPR}, pages 26296--26306, 2024{\natexlab{a}}.

\bibitem[Liu et~al.(2024{\natexlab{b}})Liu, Li, Wu, and Lee]{llava}
Haotian Liu, Chunyuan Li, Qingyang Wu, and Yong~Jae Lee.
\newblock {Visual Instruction Tuning}.
\newblock \emph{NeurIPS}, 36, 2024{\natexlab{b}}.

\bibitem[Liu et~al.(2023)Liu, Zeng, Ren, Li, Zhang, Yang, Li, Yang, Su, Zhu,
  et~al.]{groundingdino}
Shilong Liu, Zhaoyang Zeng, Tianhe Ren, Feng Li, Hao Zhang, Jie Yang, Chunyuan
  Li, Jianwei Yang, Hang Su, Jun Zhu, et~al.
\newblock {Grounding DINO: Marrying DINO with Grounded Pre-Training for
  Open-Set Object Detection}.
\newblock \emph{arXiv preprint arXiv:2303.05499}, 2023.

\bibitem[Liu et~al.(2021)Liu, Lin, Cao, Hu, Wei, Zhang, Lin, and
  Guo]{swin_transformer}
Ze Liu, Yutong Lin, Yue Cao, Han Hu, Yixuan Wei, Zheng Zhang, Stephen Lin, and
  Baining Guo.
\newblock {Swin Transformer: Hierarchical Vision Transformer Using Shifted
  Windows}.
\newblock In \emph{ICCV}, pages 10012--10022, 2021.

\bibitem[Liu et~al.(2024{\natexlab{c}})Liu, Chu, Zang, Wei, Dong, Zhang, Liang,
  Xiong, Qiao, Lin, et~al.]{mmdu}
Ziyu Liu, Tao Chu, Yuhang Zang, Xilin Wei, Xiaoyi Dong, Pan Zhang, Zijian
  Liang, Yuanjun Xiong, Yu Qiao, Dahua Lin, et~al.
\newblock {MMDU: A Multi-Turn Multi-Image Dialog Understanding Benchmark and
  Instruction-Tuning Dataset for LVLMs}.
\newblock \emph{arXiv preprint arXiv:2406.11833}, 2024{\natexlab{c}}.

\bibitem[Masry et~al.(2022)Masry, Long, Tan, Joty, and Hoque]{ChartQA}
Ahmed Masry, Do Long, Jia~Qing Tan, Shafiq Joty, and Enamul Hoque.
\newblock {ChartQA: A Benchmark for Question Answering about Charts with Visual
  and Logical Reasoning}.
\newblock In \emph{ACL}, 2022.

\bibitem[McKinzie et~al.(2024)McKinzie, Gan, Fauconnier, Dodge, Zhang, Dufter,
  Shah, Du, Peng, Weers, et~al.]{mm1}
Brandon McKinzie, Zhe Gan, Jean-Philippe Fauconnier, Sam Dodge, Bowen Zhang,
  Philipp Dufter, Dhruti Shah, Xianzhi Du, Futang Peng, Floris Weers, et~al.
\newblock {MM1: Methods, Analysis \& Insights from Multimodal LLM
  Pre-training}.
\newblock \emph{arXiv preprint arXiv:2403.09611}, 2024.

\bibitem[OpenAI(2024)]{gpt4omini}
OpenAI.
\newblock {GPT-4o mini: advancing cost-efficient intelligence}.
\newblock
  \url{https://openai.com/index/gpt-4o-mini-advancing-cost-efficient-intelligence/},
  2024.

\bibitem[Qi et~al.(2024)Qi, Ding, Wang, Bai, Lv, Hong, Xu, Hou, Li, Dong,
  et~al.]{cogcom}
Ji Qi, Ming Ding, Weihan Wang, Yushi Bai, Qingsong Lv, Wenyi Hong, Bin Xu, Lei
  Hou, Juanzi Li, Yuxiao Dong, et~al.
\newblock {CogCoM: Train Large Vision-Language Models Diving into Details
  through Chain of Manipulations}.
\newblock \emph{arXiv preprint arXiv:2402.04236}, 2024.

\bibitem[Radford et~al.(2021{\natexlab{a}})Radford, Kim, Hallacy, Ramesh, Goh,
  Agarwal, Sastry, Askell, Mishkin, Clark, et~al.]{clip}
Alec Radford, Jong~Wook Kim, Chris Hallacy, Aditya Ramesh, Gabriel Goh,
  Sandhini Agarwal, Girish Sastry, Amanda Askell, Pamela Mishkin, Jack Clark,
  et~al.
\newblock {Learning Transferable Visual Models From Natural Language
  Supervision}.
\newblock In \emph{ICML}, pages 8748--8763. PMLR, 2021{\natexlab{a}}.

\bibitem[Radford et~al.(2021{\natexlab{b}})Radford, Kim, Hallacy, Ramesh, Goh,
  Agarwal, Sastry, Askell, Mishkin, Clark, et~al.]{vit}
Alec Radford, Jong~Wook Kim, Chris Hallacy, Aditya Ramesh, Gabriel Goh,
  Sandhini Agarwal, Girish Sastry, Amanda Askell, Pamela Mishkin, Jack Clark,
  et~al.
\newblock {Learning Transferable Visual Models From Natural Language
  Supervision}.
\newblock In \emph{ICML}, pages 8748--8763. PMLR, 2021{\natexlab{b}}.

\bibitem[Reid et~al.(2024)Reid, Savinov, Teplyashin, Lepikhin, Lillicrap,
  Alayrac, Soricut, Lazaridou, Firat, Schrittwieser, et~al.]{gemini}
Machel Reid, Nikolay Savinov, Denis Teplyashin, Dmitry Lepikhin, Timothy
  Lillicrap, Jean-baptiste Alayrac, Radu Soricut, Angeliki Lazaridou, Orhan
  Firat, Julian Schrittwieser, et~al.
\newblock {Gemini 1.5: Unlocking multimodal understanding across millions of
  tokens of context}.
\newblock \emph{arXiv preprint arXiv:2403.05530}, 2024.

\bibitem[Schuhmann et~al.(2022)Schuhmann, Beaumont, Vencu, Gordon, Wightman,
  Cherti, Coombes, Katta, Mullis, Wortsman, et~al.]{laion5b}
Christoph Schuhmann, Romain Beaumont, Richard Vencu, Cade Gordon, Ross
  Wightman, Mehdi Cherti, Theo Coombes, Aarush Katta, Clayton Mullis, Mitchell
  Wortsman, et~al.
\newblock {LAION-5B: An open large-scale dataset for training next generation
  image-text models}.
\newblock \emph{NeurIPS}, 35:\penalty0 25278--25294, 2022.

\bibitem[Seo et~al.(2017)Seo, Lehrmann, Han, and Sigal]{mnist-dialog}
Paul~Hongsuck Seo, Andreas Lehrmann, Bohyung Han, and Leonid Sigal.
\newblock {Visual Reference Resolution using Attention Memory for Visual
  Dialog}.
\newblock \emph{NeurIPS}, 30, 2017.

\bibitem[Shao et~al.(2024)Shao, Qian, Xiao, Song, Zong, Wang, Liu, and
  Li]{viscot}
Hao Shao, Shengju Qian, Han Xiao, Guanglu Song, Zhuofan Zong, Letian Wang, Yu
  Liu, and Hongsheng Li.
\newblock {Visual CoT: Advancing Multi-Modal Language Models with a
  Comprehensive Dataset and Benchmark for Chain-of-Thought Reasoning}.
\newblock \emph{arXiv preprint arXiv:2403.16999}, 2024.

\bibitem[Shuster et~al.(2018)Shuster, Humeau, Bordes, and Weston]{IGC}
Kurt Shuster, Samuel Humeau, Antoine Bordes, and Jason Weston.
\newblock {Image Chat: Engaging Grounded Conversations}.
\newblock \emph{arXiv preprint arXiv:1811.00945}, 2018.

\bibitem[Singh et~al.(2019)Singh, Natarjan, Shah, Jiang, Chen, Parikh, and
  Rohrbach]{TextVQA}
Amanpreet Singh, Vivek Natarjan, Meet Shah, Yu Jiang, Xinlei Chen, Devi Parikh,
  and Marcus Rohrbach.
\newblock {Towards VQA Models That Can Read}.
\newblock In \emph{CVPR}, pages 8317--8326, 2019.

\bibitem[Stureborg et~al.(2024)Stureborg, Alikaniotis, and Suhara]{llm_eval_2}
Rickard Stureborg, Dimitris Alikaniotis, and Yoshi Suhara.
\newblock {Large language models are inconsistent and biased evaluators}.
\newblock \emph{arXiv preprint arXiv:2405.01724}, 2024.

\bibitem[Tan et~al.(2024)Tan, Ding, Zhang, Li, Zhou, Yue, Xia, Jiang, Zheng,
  Xu, et~al.]{tan2024towards}
Weihao Tan, Ziluo Ding, Wentao Zhang, Boyu Li, Bohan Zhou, Junpeng Yue,
  Haochong Xia, Jiechuan Jiang, Longtao Zheng, Xinrun Xu, et~al.
\newblock {Towards General Computer Control: A Multimodal Agent for Red Dead
  Redemption II as a Case Study}.
\newblock In \emph{ICLR 2024 Workshop on Large Language Model (LLM) Agents},
  2024.

\bibitem[Tian et~al.(2024)Tian, Ma, Xie, Qiu, Tang, Zhang, Jiao, Tian, and
  Ye]{chatterbox}
Yunjie Tian, Tianren Ma, Lingxi Xie, Jihao Qiu, Xi Tang, Yuan Zhang, Jianbin
  Jiao, Qi Tian, and Qixiang Ye.
\newblock {ChatterBox: Multi-round Multimodal Referring and Grounding}.
\newblock \emph{arXiv preprint arXiv:2401.13307}, 2024.

\bibitem[Touvron et~al.(2023)Touvron, Lavril, Izacard, Martinet, Lachaux,
  Lacroix, Rozi{\`e}re, Goyal, Hambro, Azhar, et~al.]{llama}
Hugo Touvron, Thibaut Lavril, Gautier Izacard, Xavier Martinet, Marie-Anne
  Lachaux, Timoth{\'e}e Lacroix, Baptiste Rozi{\`e}re, Naman Goyal, Eric
  Hambro, Faisal Azhar, et~al.
\newblock {LLaMA: Open and Efficient Foundation Language Models}.
\newblock \emph{arXiv preprint arXiv:2302.13971}, 2023.

\bibitem[Towers et~al.(2024)Towers, Kwiatkowski, Terry, Balis, De~Cola, Deleu,
  Goul{\~a}o, Kallinteris, Krimmel, KG, et~al.]{gymnasium}
Mark Towers, Ariel Kwiatkowski, Jordan Terry, John~U Balis, Gianluca De~Cola,
  Tristan Deleu, Manuel Goul{\~a}o, Andreas Kallinteris, Markus Krimmel, Arjun
  KG, et~al.
\newblock {Gymnasium: A Standard Interface for Reinforcement Learning
  Environments}.
\newblock \emph{arXiv preprint arXiv:2407.17032}, 2024.

\bibitem[Vaswani et~al.(2017)Vaswani, Shazeer, Parmar, Uszkoreit, Jones, Gomez,
  Kaiser, and Polosukhin]{transformer}
Ashish Vaswani, Noam Shazeer, Niki Parmar, Jakob Uszkoreit, Llion Jones,
  Aidan~N Gomez, {\L}ukasz Kaiser, and Illia Polosukhin.
\newblock {Attention Is All You Need}.
\newblock \emph{NeurIPS}, 30, 2017.

\bibitem[Wei et~al.(2022)Wei, Wang, Schuurmans, Bosma, Xia, Chi, Le, Zhou,
  et~al.]{cot}
Jason Wei, Xuezhi Wang, Dale Schuurmans, Maarten Bosma, Fei Xia, Ed Chi, Quoc~V
  Le, Denny Zhou, et~al.
\newblock {Chain-of-Thought Prompting Elicits Reasoning in Large Language
  Models}.
\newblock \emph{NeurIPS}, 35:\penalty0 24824--24837, 2022.

\bibitem[Xu et~al.(2024)Xu, Wang, Xu, Ding, Jiang, Ding, and
  Karlsson]{game-mllm-survey}
Xinrun Xu, Yuxin Wang, Chaoyi Xu, Ziluo Ding, Jiechuan Jiang, Zhiming Ding, and
  Börje~F. Karlsson.
\newblock {A Survey on Game Playing Agents and Large Models: Methods,
  Applications, and Challenges}.
\newblock \emph{arXiv preprint arXiv:2403.10249}, 2024.

\bibitem[Yin et~al.(2023)Yin, Fu, Zhao, Li, Sun, Xu, and Chen]{mllmsurvey}
Shukang Yin, Chaoyou Fu, Sirui Zhao, Ke Li, Xing Sun, Tong Xu, and Enhong Chen.
\newblock {A Survey on Multimodal Large Language Models}.
\newblock \emph{arXiv preprint arXiv:2306.13549}, 2023.

\bibitem[Zheng et~al.(2024)Zheng, Liu, Feng, and Lu]{steve-eye}
Sipeng Zheng, Jiazheng Liu, Yicheng Feng, and Zongqing Lu.
\newblock {Steve-Eye: Equipping LLM-based Embodied Agents with Visual
  Perception in Open Worlds}.
\newblock In \emph{ICLR}, 2024.

\bibitem[Zheng et~al.(2025)Zheng, Zhou, Feng, Wang, and Lu]{zheng2025unicode}
Sipeng Zheng, Bohan Zhou, Yicheng Feng, Ye Wang, and Zongqing Lu.
\newblock {UniCode: Learning a Unified Codebook for Multimodal Large Language
  Models}.
\newblock In \emph{ECCV}, pages 426--443. Springer, 2025.

\end{thebibliography}
}

\newpage
\setcounter{section}{0}
\renewcommand{\thesection}{\Alph{section}}


\clearpage
\section{Dataset}

We use GPT-4o-mini~\cite{gpt4omini} to generate our \Dataset\ dataset. Our dataset mainly consists of three parts: everyday scenes, tabular scenes, and Minigrid settings. 
We adopt different prompts for the generation of datasets under different scenes. 

\subsection{Dataset Collection}

We design prompts for different scenarios, and the same devising ideas can be used in other scenarios for data collection. 

\begin{figure}[ht]
  \centering
  \includegraphics[width=1\linewidth]{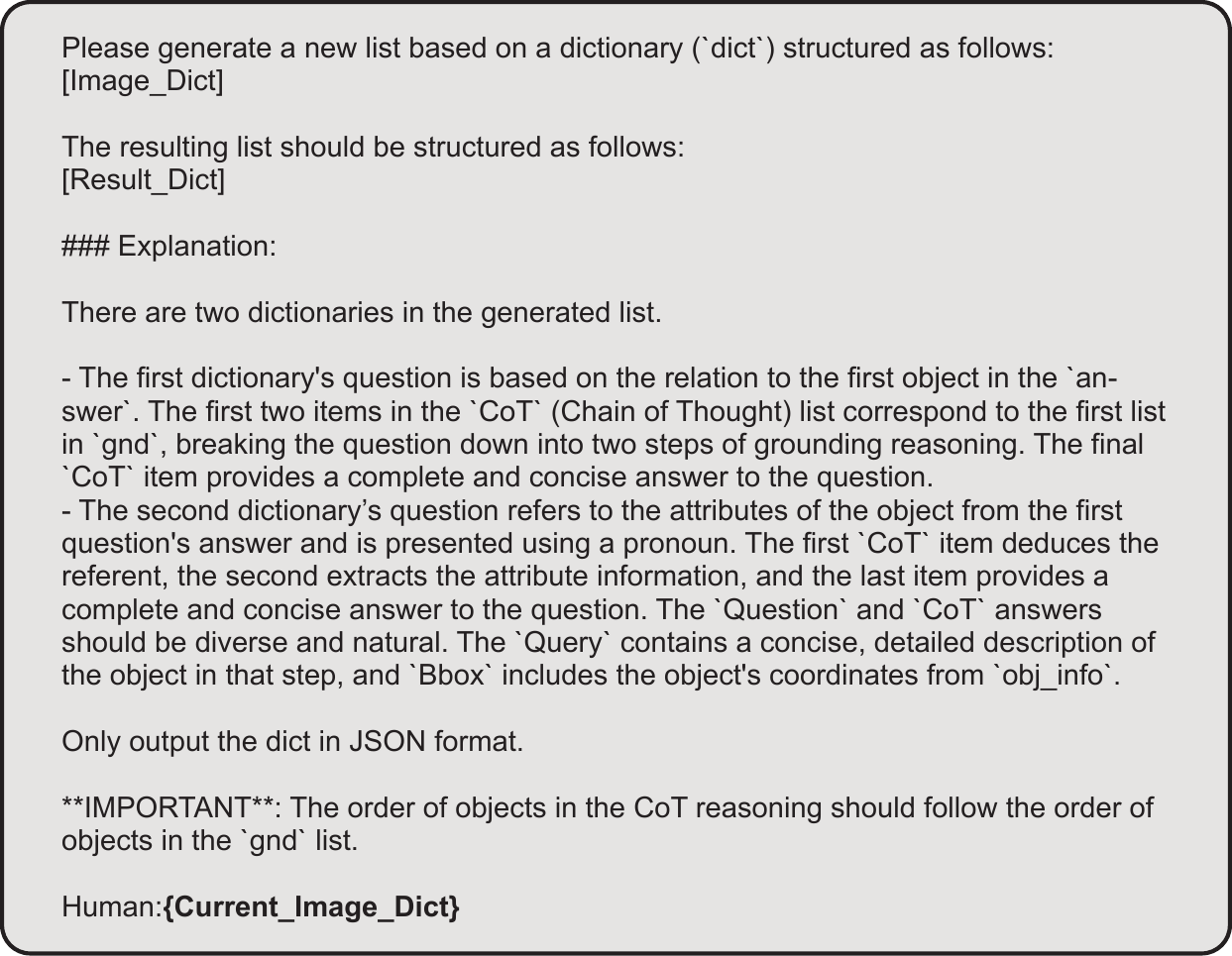}
   \caption{The first example prompt for generating data samples in everyday scenes.}
   \label{fig:prompt_vg_1}
\end{figure}

\begin{figure}[ht]
  \centering
  \includegraphics[width=1\linewidth]{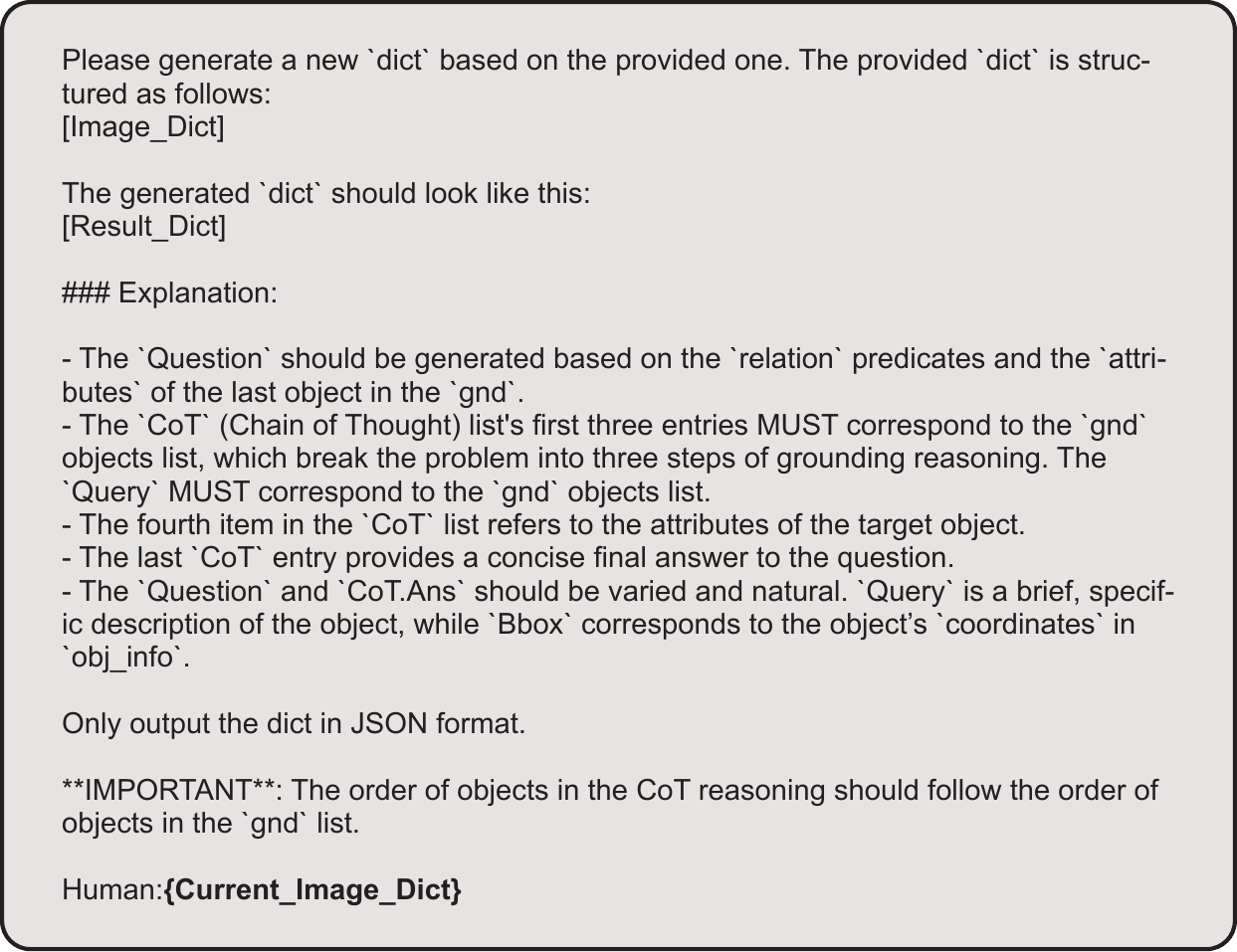}
   \caption{The second example prompt for generating data samples in everyday scenes.}
   \label{fig:prompt_vg_2}
\end{figure}

\begin{figure}[ht]
  \centering
  \includegraphics[width=1\linewidth]{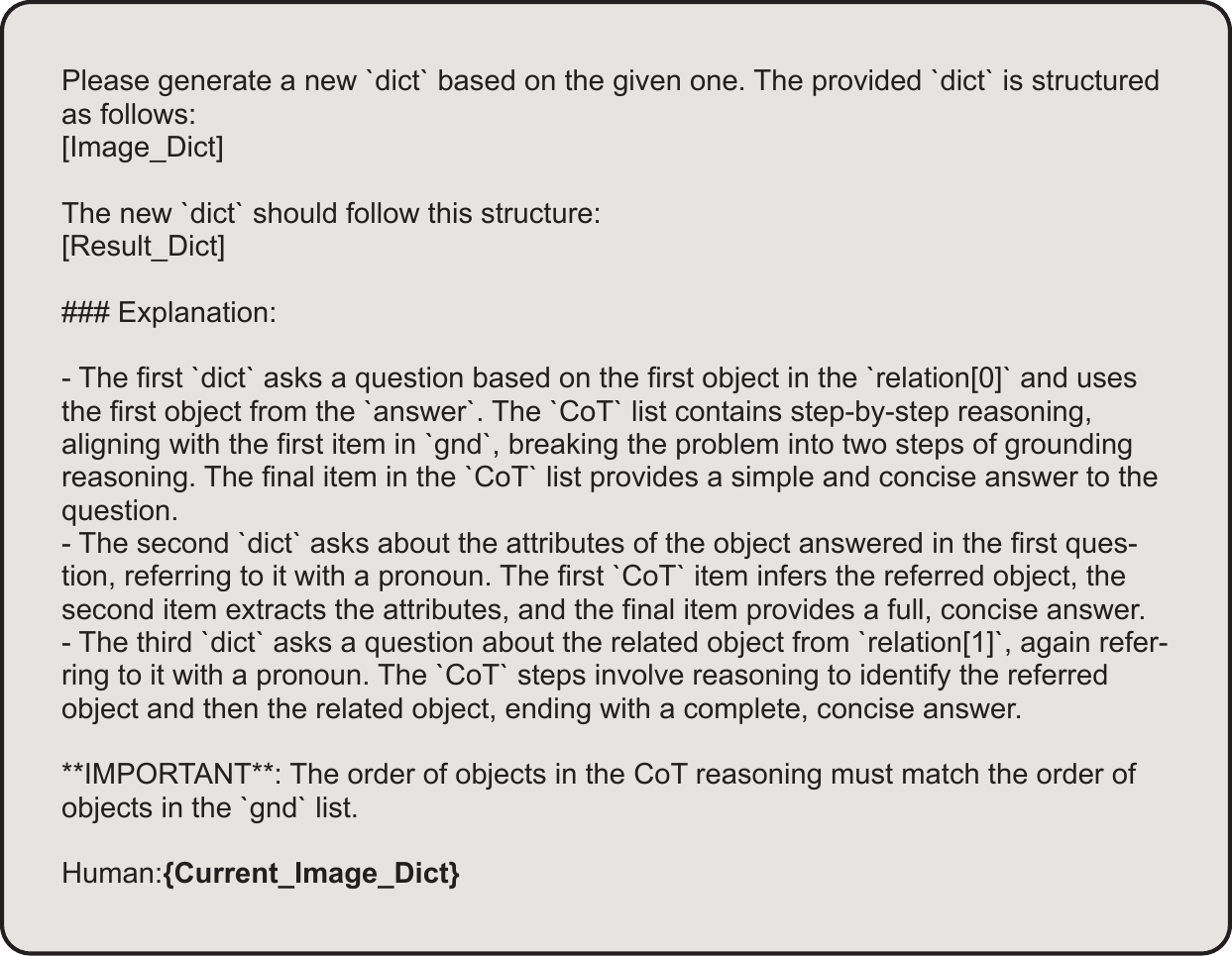}
   \caption{The third example prompt for generating data samples in everyday scenes.}
   \label{fig:prompt_vg_3}
\end{figure}

\noindent \textbf{Everyday Scenes.}
\label{Appendix:everyday_scene}
For everyday scenes, we generate our dataset from the Visual Genome dataset~\cite{Visual_Genome}. 
Since the original dataset has human-annotated attributes and relationship data, we extract the subsets that represent the QA pairs and feed them to GPT-4o-mini to generate corresponding dialogues. \Cref{fig:prompt_vg_1,fig:prompt_vg_2,fig:prompt_vg_3} show several example prompts. 

\begin{figure}[ht]
  \centering
  \includegraphics[width=1\linewidth]{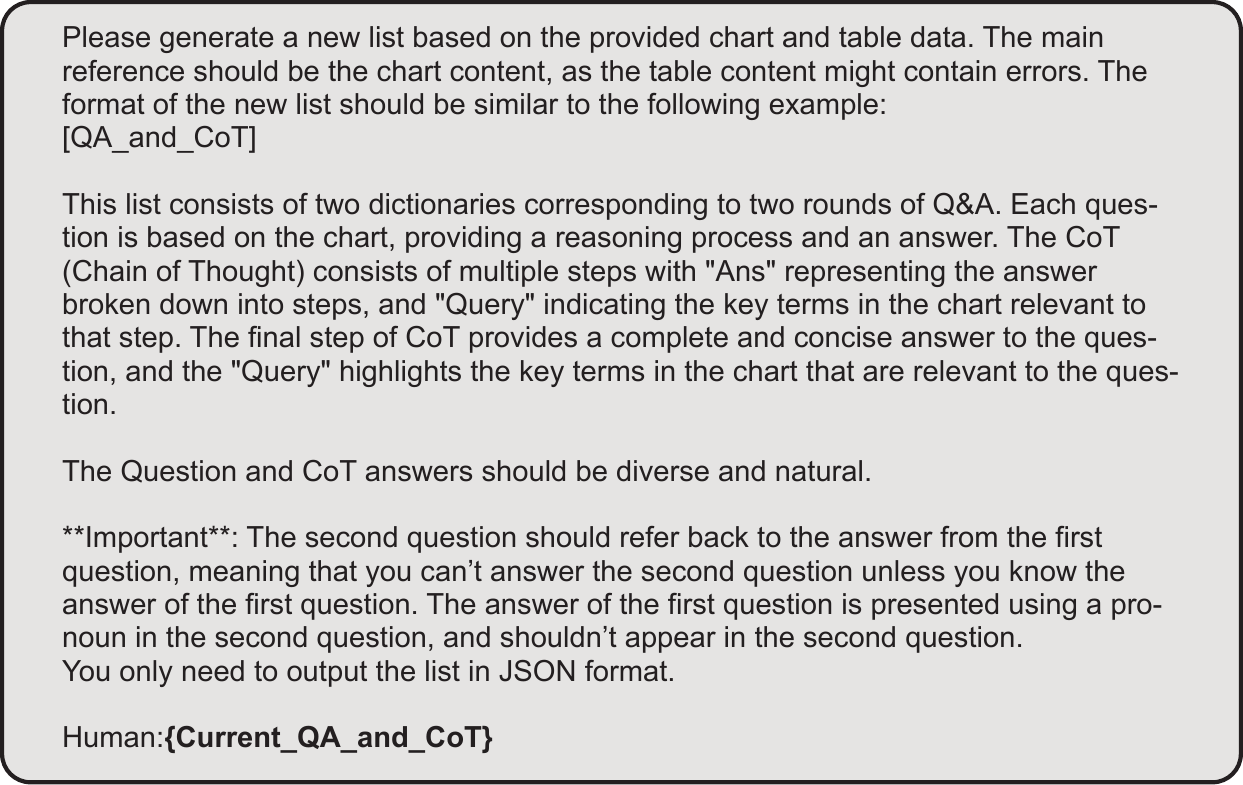}
   \caption{The prompt structure to generate samples in tabular scenes.}
   \label{fig:prompt_chart_0}
\end{figure}

\begin{figure}[ht]
  \centering
  \includegraphics[width=1\linewidth]{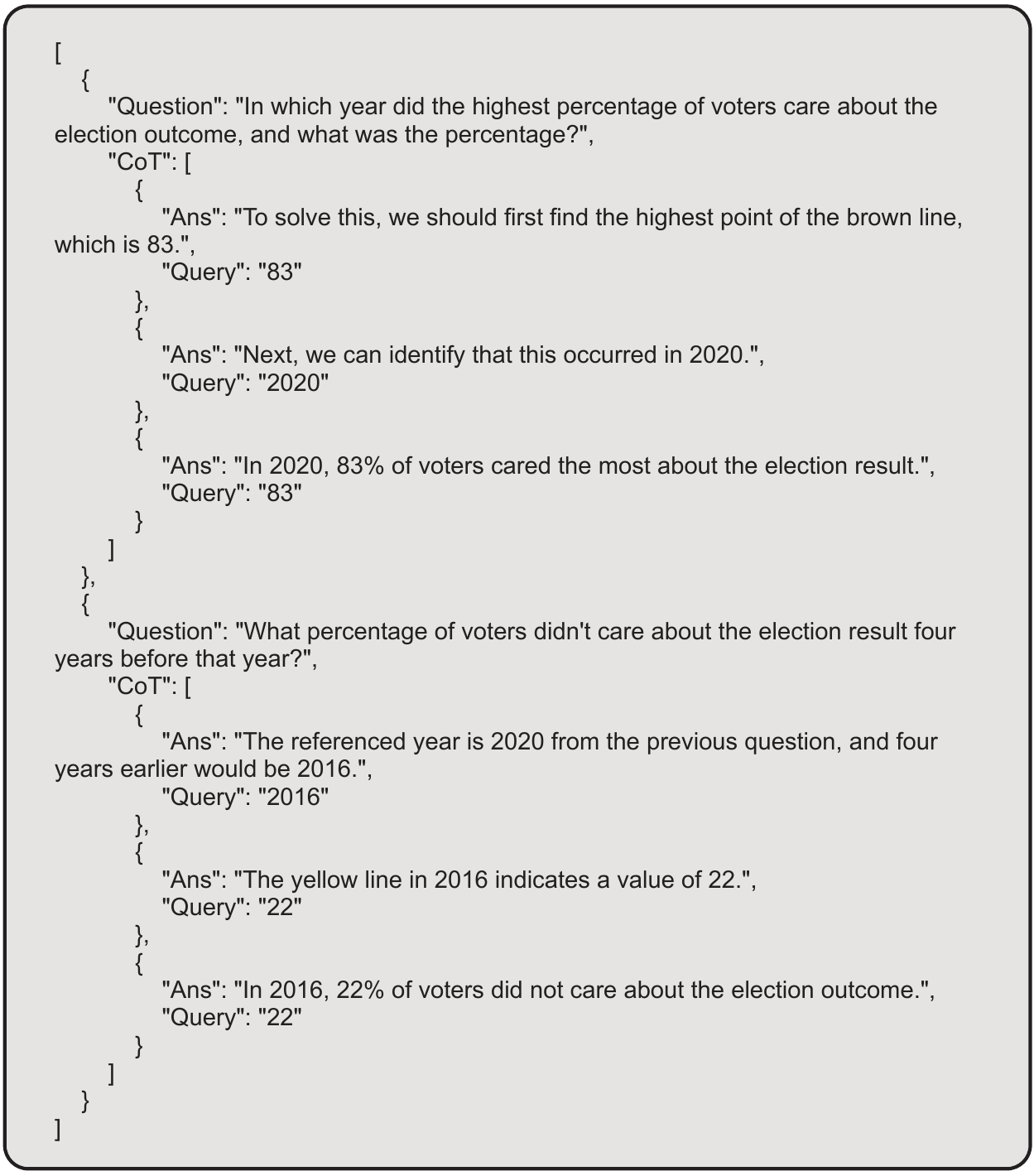}
   \caption{The question-answer (QA) and Chain-of-Thought (CoT) examples for line charts.}
   \label{fig:prompt_chart_1}
\end{figure}

\begin{figure}[ht]
  \centering
  \includegraphics[width=1\linewidth]{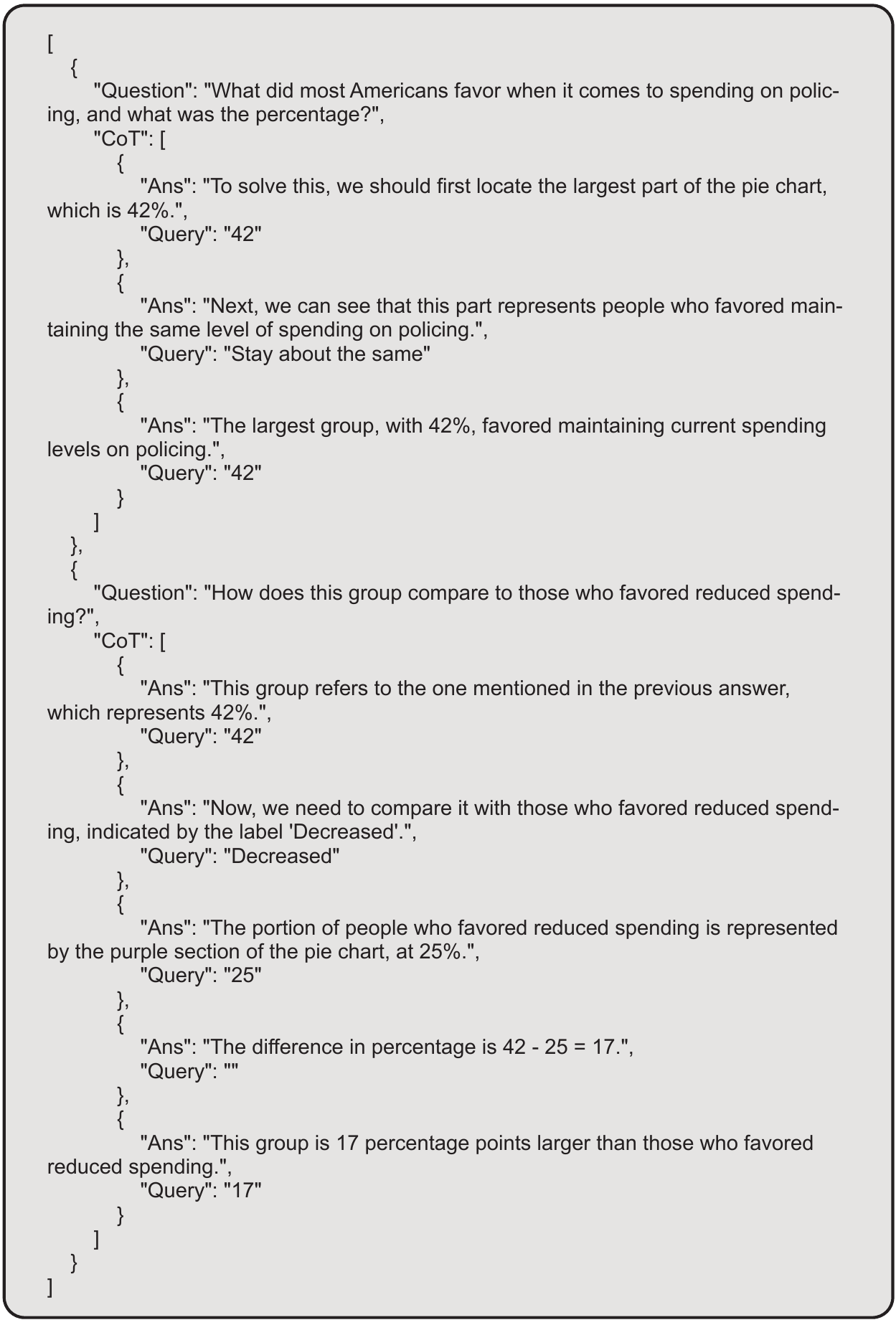}
   \caption{The question-answer (QA) and Chain-of-Thought (CoT) examples for pie charts.}
   \label{fig:prompt_chart_2}
\end{figure}

\begin{figure}[ht]
  \centering
  \includegraphics[width=1\linewidth]{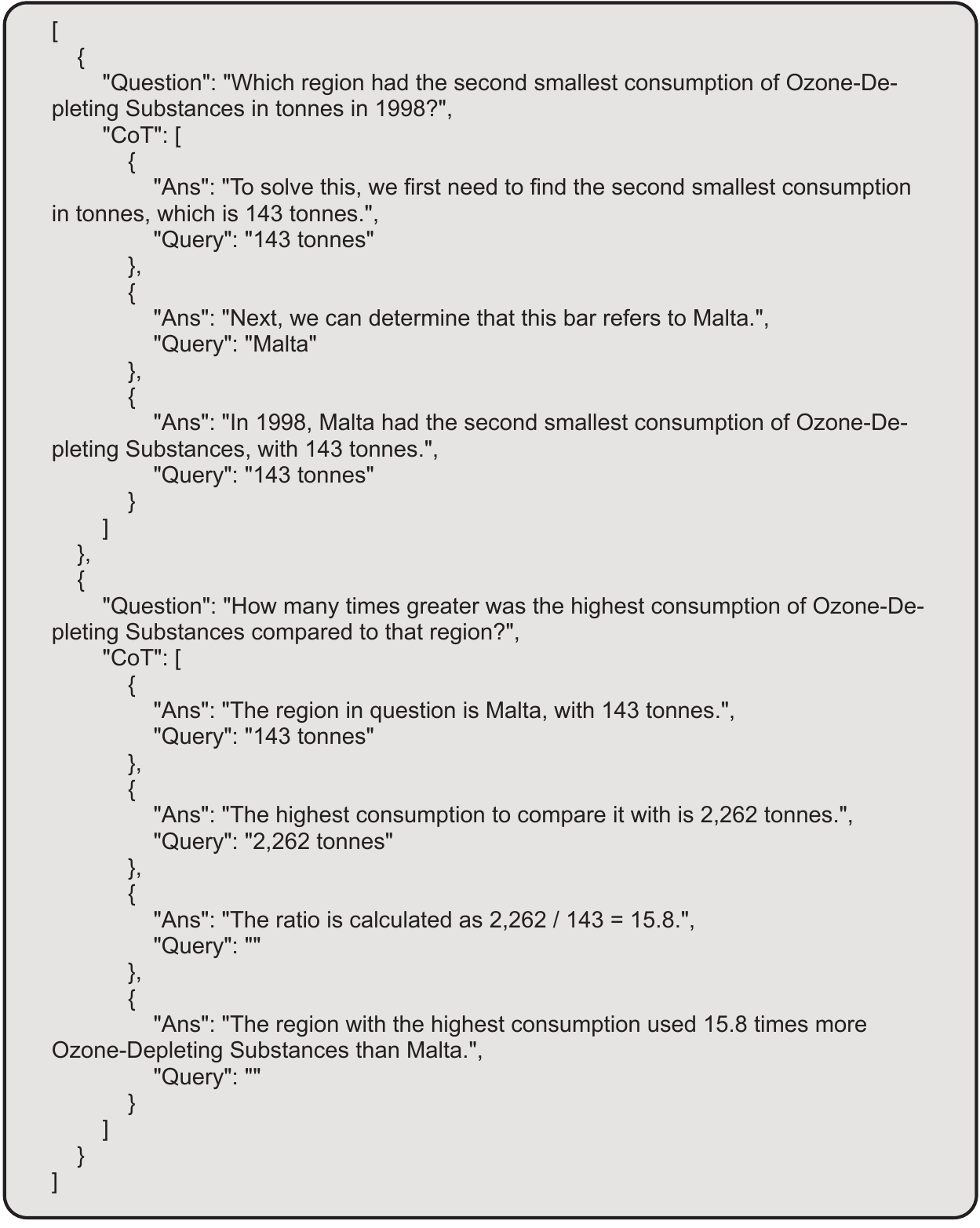}
   \caption{The question-answer (QA) and Chain-of-Thought (CoT) examples for bar charts.}
   \label{fig:prompt_chart_3}
\end{figure}

\noindent \textbf{Tabular Scenes.}
\label{Appendix:tabular_scene}
For tabular scenes, we generate our dataset from the ChartQA dataset~\cite{ChartQA}. In general, we use different types of graphs to capture various visualization intuitions, providing corresponding chart examples in the prompts.
\Cref{fig:prompt_chart_0} illustrates the main structure of the prompt, while \Cref{fig:prompt_chart_1,fig:prompt_chart_2,fig:prompt_chart_3} show examples for line, pie, and bar charts, respectively.

\begin{figure}[ht]
  \centering
  \includegraphics[width=1\linewidth]{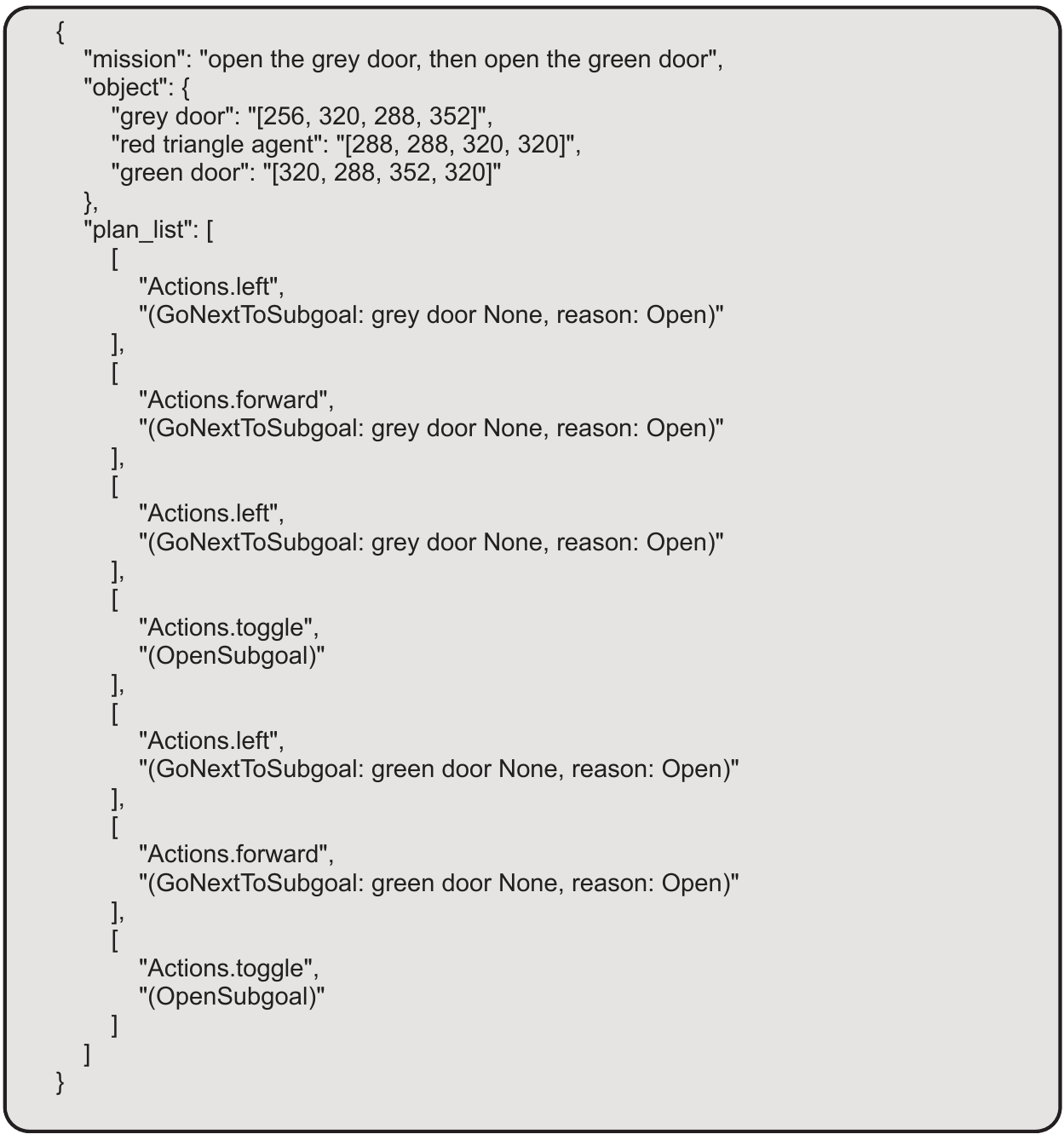}
   \caption{The mission and plan input example of Minigrid settings.}
   \label{fig:prompt_minigrid_1}
\end{figure}

\begin{figure}[ht]
  \centering
  \includegraphics[width=1\linewidth]{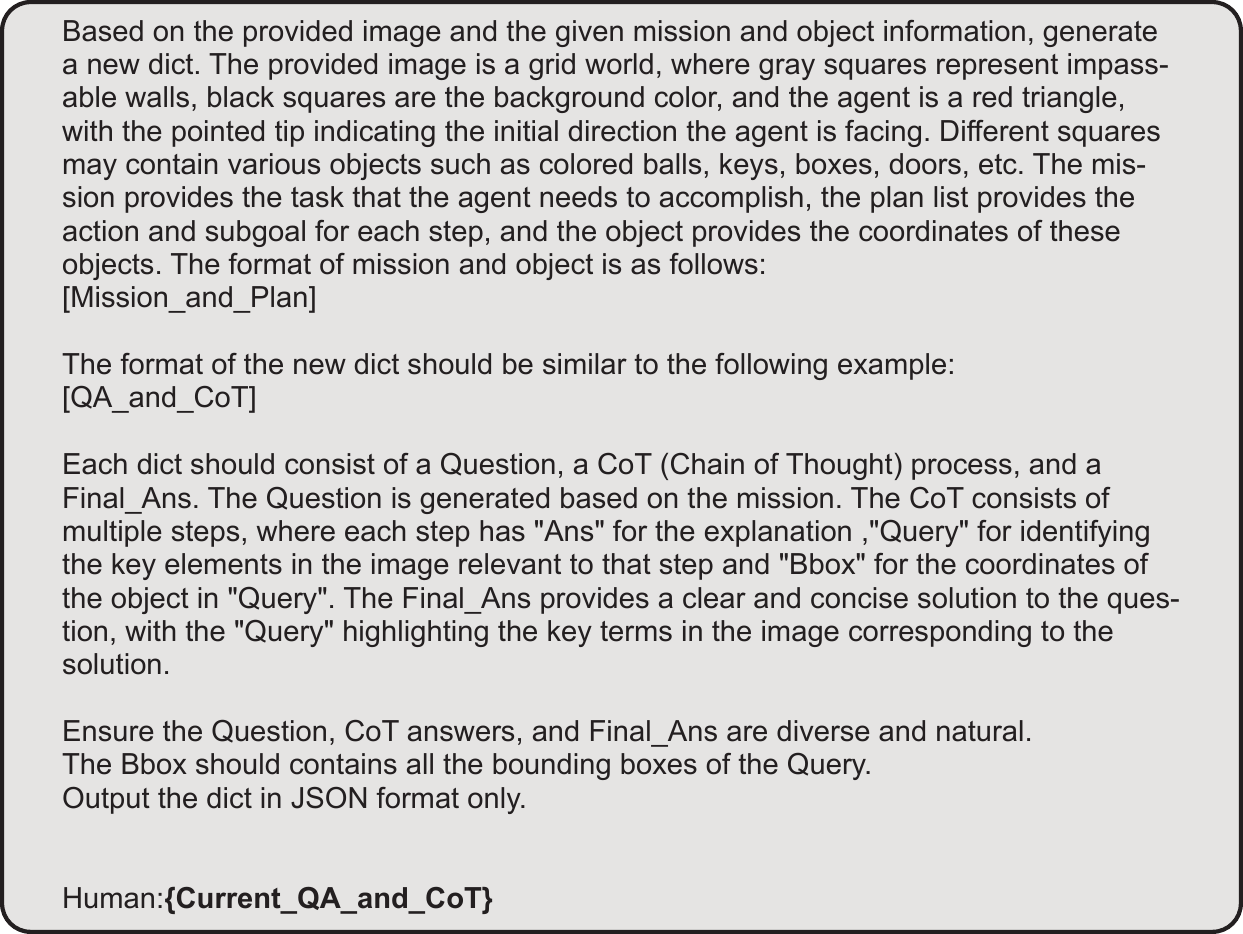}
   \caption{The prompt structure to generate data samples in Minigrid settings.}
   \label{fig:prompt_minigrid_0}
\end{figure}

\noindent \textbf{Minigrid Settings.}
\label{Appendix:minigrid}
For Minigrid settings, we generate our dataset from the Minigrid database~\cite{Minigrid}. 
Since we observe that GPT-4o-mini struggles to solve the mission without ground-truth planning, we first use BabyAI~\cite{babyai} to collect the plan needed to complete the mission for each environment generated by the Minigrid database.
We then combine the positions of all objects with the mission and plan, as shown in \Cref{fig:prompt_minigrid_1}, and feed them to GPT-4o-mini. 
The prompt structure is illustrated in \Cref{fig:prompt_minigrid_0}.

\subsection{Dataset Format}
\label{Appendix:dataset_format}

\begin{figure}[ht]
	\centering
	\begin{subfigure}{1\linewidth}
		\centering
		\includegraphics[width=1\linewidth]{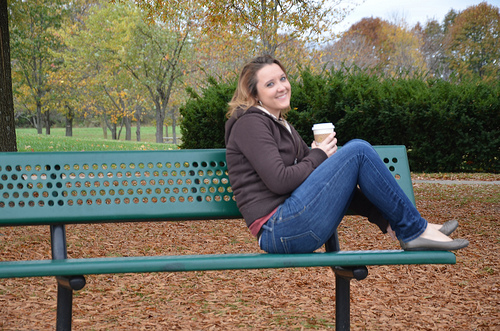}
		\caption{the original image}
		\label{subfig:format_vg_img}
	\end{subfigure}
    
	\vspace{2mm}
    
	\begin{subfigure}{1\linewidth}
		\centering
		\includegraphics[width=1\linewidth]{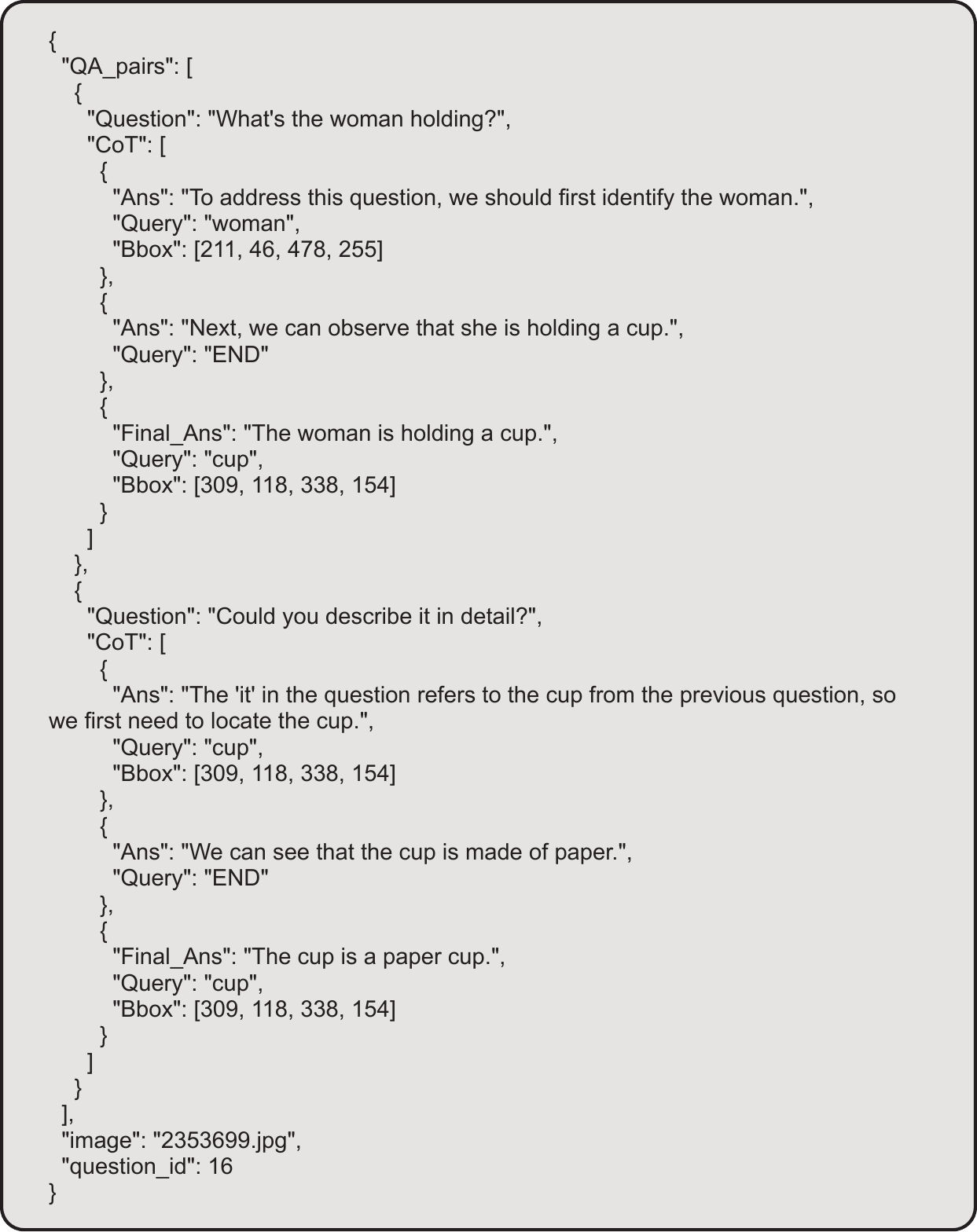}
		\caption{the sample format}
		\label{subfig:format_vg_json}
	\end{subfigure}
	\centering
	\caption{One example of the original image and the generated sample from Visual Genome in JSON format. }
	\label{fig:data_format_vg}
\end{figure}

\begin{figure}[ht]
	\centering
	\begin{subfigure}{1\linewidth}
		\centering
		\includegraphics[width=1\linewidth]{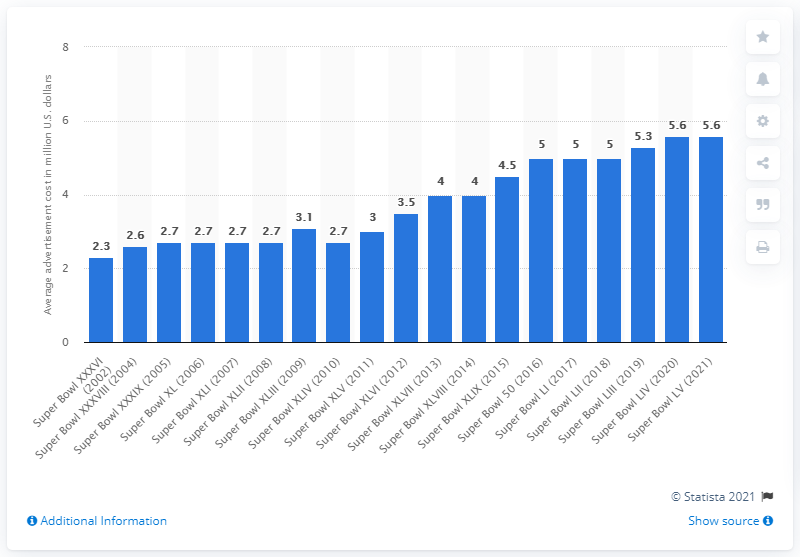}
		\caption{the original image}
		\label{subfig:format_chart_img}
	\end{subfigure}
	    
	\vspace{2mm}
    
	\begin{subfigure}{1\linewidth}
		\centering
		\includegraphics[width=1\linewidth]{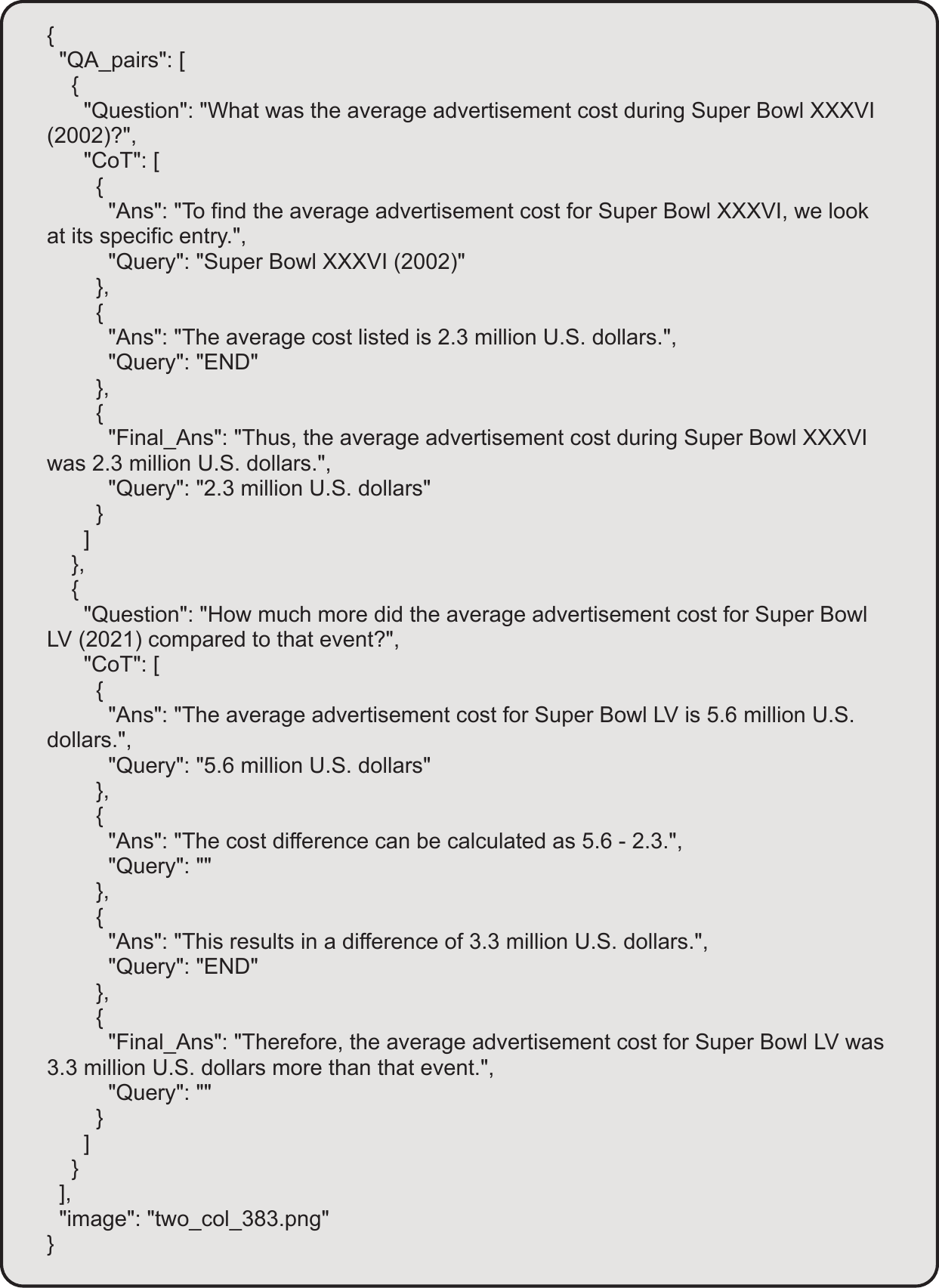}
		\caption{the sample format}
		\label{subfig:format_chart_json}
	\end{subfigure}
	\centering
	\caption{One example of the original image and the generated data point from ChartQA in JSON format. The bounding boxes of the queries are generated using EasyOCR~\cite{easyocr} and thus are not shown in the example.}
	\label{fig:data_format_chart}
\end{figure}

\begin{figure}[ht]
	\centering
	\begin{subfigure}{1\linewidth}
		\centering
		\includegraphics[width=1\linewidth]{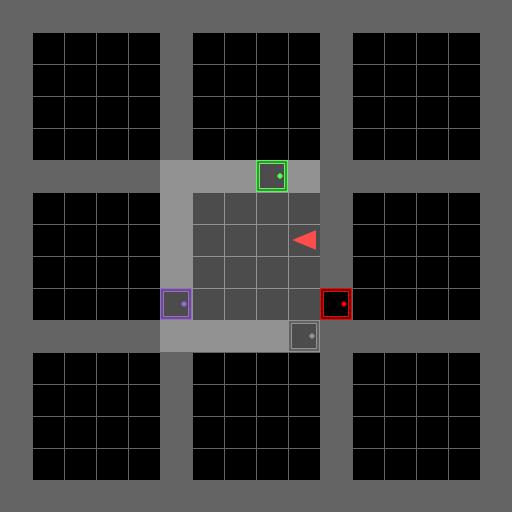}
		\caption{the original image}
		\label{subfig:format_babyai_img}
	\end{subfigure}
		    
	\vspace{2mm}
    
	\begin{subfigure}{1\linewidth}
		\centering
		\includegraphics[width=1\linewidth]{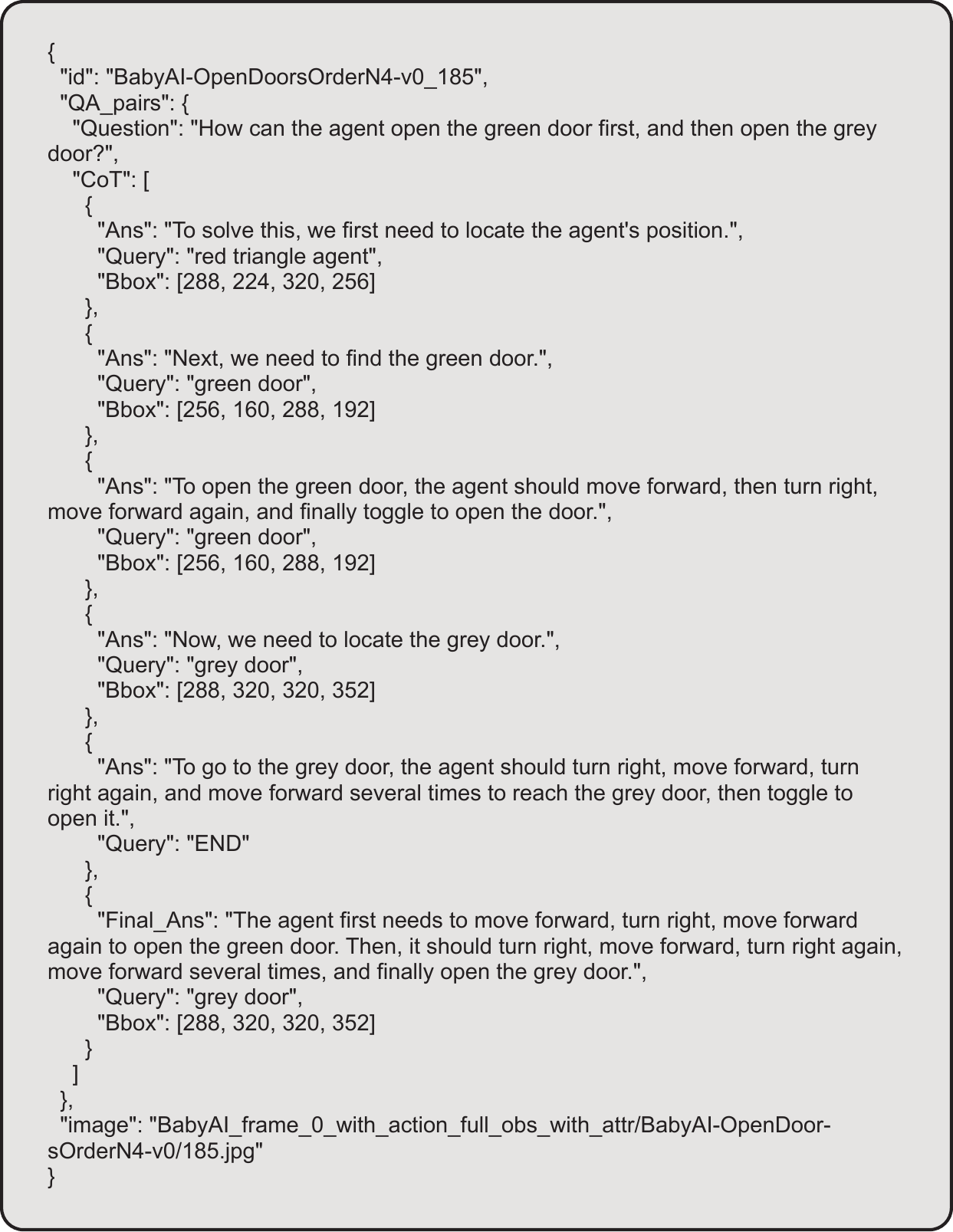}
		\caption{the sample format}
		\label{subfig:format_babyai_json}
	\end{subfigure}
	\centering
	\caption{One example of the original image and the generated sample from Minigrid in JSON format.}
	\label{fig:data_format_babyai}
\end{figure}

Examples of the final \Dataset\ dataset are shown in \Cref{fig:data_format_vg,fig:data_format_chart,fig:data_format_babyai}. \Cref{subfig:format_vg_img,subfig:format_chart_img,subfig:format_babyai_img} display the original images from the source datasets and environments, while \Cref{subfig:format_vg_json,subfig:format_chart_json,subfig:format_babyai_json} show the data format of \Dataset\ generated by GPT-4o-mini and standardized according to specific rules.

\subsection{Evaluation}
\label{Appendix:evaluation}

\begin{figure}[h]
  \centering
  \includegraphics[width=1\linewidth]{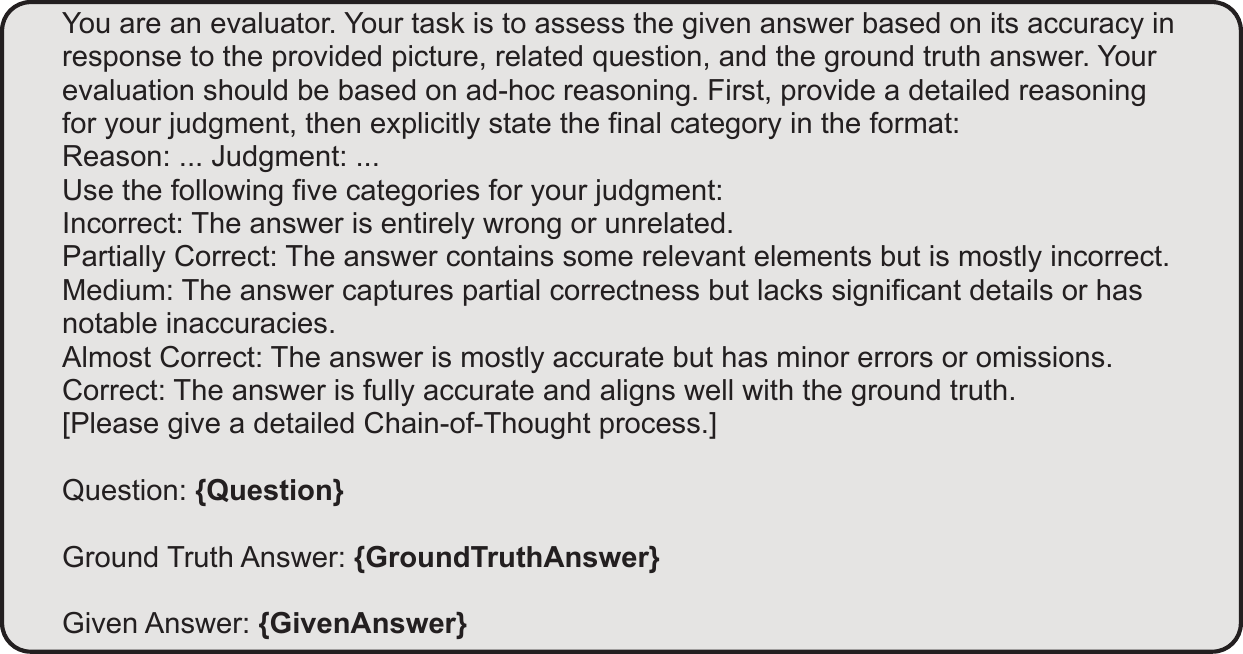}
   \caption{The evaluation prompt structure given to Gemini-1.5-Pro. The content in `[]' is added when the CoT process is evaluated. }
   \label{fig:prompt_azure_ans}
\end{figure}

Since GPT-4o-mini contributes to generating our datasets, we use Gemini-1.5-Pro~\cite{gemini} for evaluation. 
There are multiple reasons for choosing it for this task: answer formatting and the Chain of Thought (CoT) processes may be diverse, making a simple similarity score insufficient for evaluation. 
Additionally, recent works~\cite{llava,steve-eye} commonly apply LLMs for judgment. 
We provide the MLLM with images, ground-truth answers, and generated responses, and ask it to score the accuracy of the generated answers across five categories.
We notice that the MLLM provides more reasonable rankings when asked to explain the `ad-hoc' reason before their final score.
As a result, we include this reasoning step in the prompt, as shown in \Cref{fig:prompt_azure_ans}.

\begin{table}[h]
\centering
\scalebox{1}{
\begin{tabular}{l|l}
\toprule
 \textbf{hyper-parameters} &
  \textbf{value} \\ \midrule
   deepspeed &
  zero3 \\ 
   base model &
  LLaVA-1.5-7B \\ 
   conversation template &
  Vicuna v1 \\ 
   vision tower &
  CLIP-ViT-Large- \\
  &  \qquad Patch14-336 \\ 
   modality projector type &
  mlp2x\_gelu \\ 
   image aspect ratio &
  pad \\ 
   training epochs &
  1 \\ 
   training batch size &
  16 \\ 
   learning rate &
  2e-5 \\ 
   weight decay &
  0 \\ 
   warm-up ratio &
  0.03 \\ 
   model max length &
  2048 \\ 
   data loader workers &
  4 \\ \bottomrule
\end{tabular}
}
\caption{The implementation details of the \Agent\ module.}
\label{tab:cerebration_param}
\end{table}

\begin{table}[h]
\centering
\scalebox{1}{
\begin{tabular}{l|l}
\toprule
 \textbf{hyper-parameters} &
  \textbf{value} \\ \midrule
   deepspeed &
  zero3 \\ 
   base model &
  LLaVA-1.5-7B \\ 
   conversation template &
  Vicuna v1 \\ 
   vision tower & CLIP-ViT-Large- \\
    &  \qquad Patch14-336 \\ 
   modality projector type &
  mlp2x\_gelu \\ 
   layer selected for & -2 \\
   \qquad fine-tuning vision tower &  \\ 
   image aspect ratio &
  pad \\ 
   training epochs &
  1 \\ 
   training batch size &
  32 \\ 
   learning rate &
  2e-5 \\ 
   weight decay &
  0 \\ 
   warm-up ratio &
  0.03 \\ 
   model max length &
  2048 \\ 
   data loader workers &
  4 \\ 
   fine-tune vision tower &
  True/False \\ \bottomrule
\end{tabular}
}
\caption{The implementation details of the \Grounder\ module.}
\label{tab:gazing_param}
\end{table}


\begin{figure*}[h]
	\centering
	\begin{subfigure}{0.45\linewidth}
		\centering
		\includegraphics[width=1\linewidth]{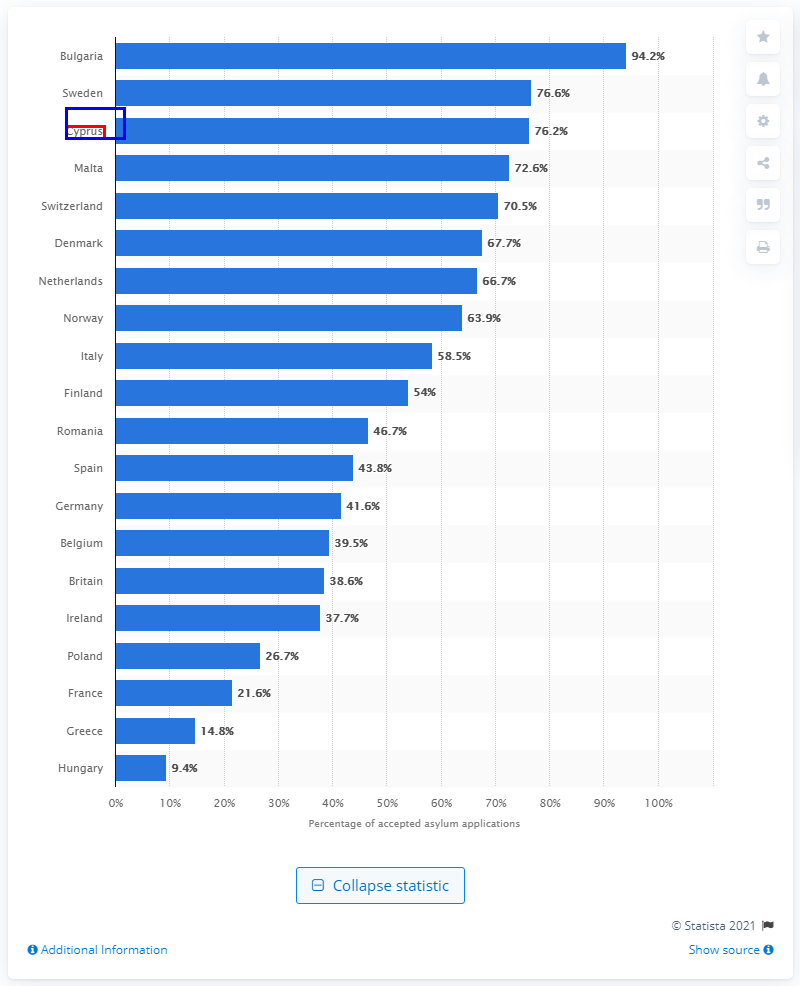}
		\caption{\Model\ }
		\label{subfig:Model_gazing_chart}
	\end{subfigure}
	\centering
	\begin{subfigure}{0.45\linewidth}
		\centering
		\includegraphics[width=1\linewidth]{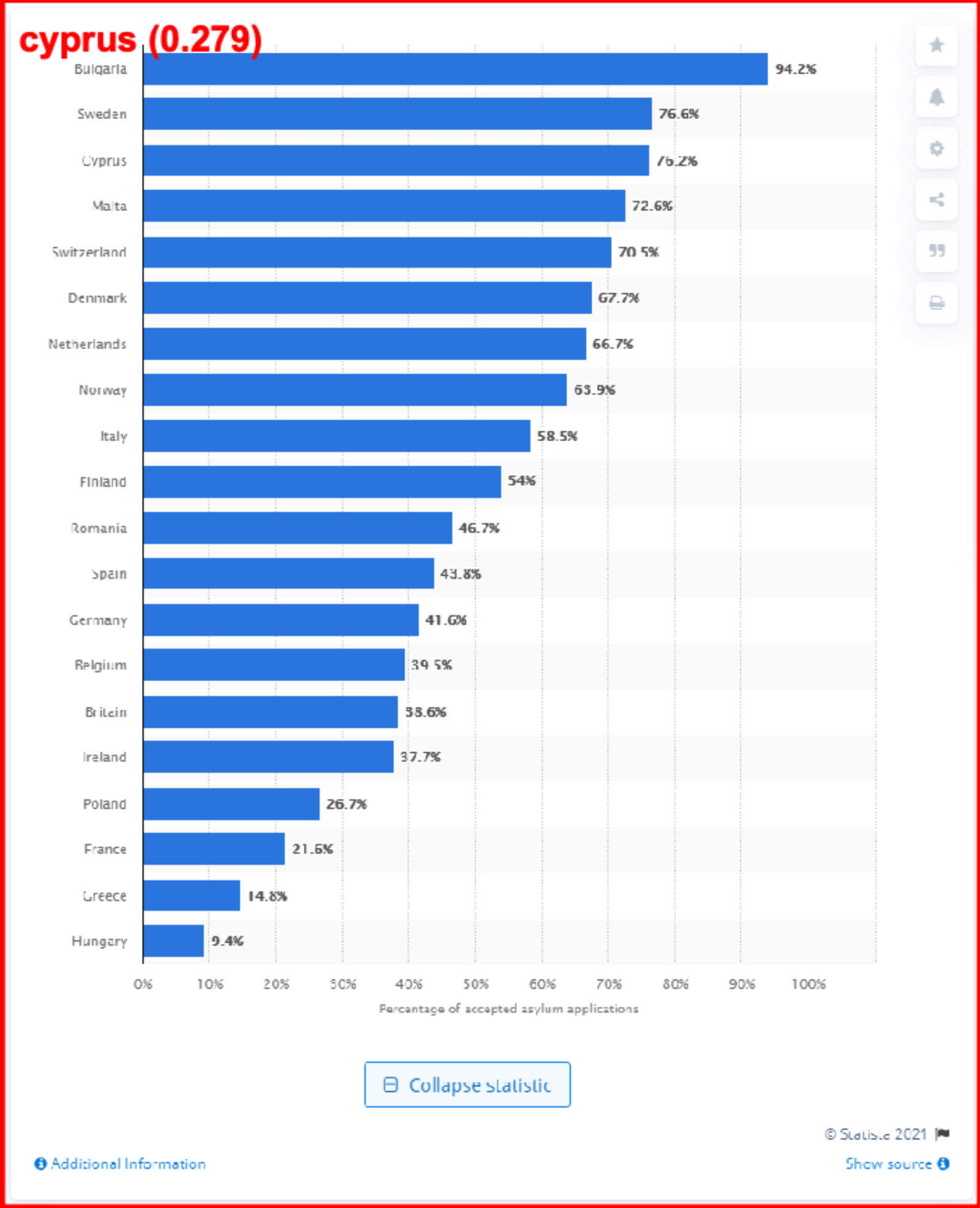}
		\caption{Grounding DINO}
		\label{subfig:DINO_gazing_chart}
	\end{subfigure}
	\centering
	\caption{The grounding comparison between Grounding DINO and the \Grounder\ module of \Model\ in Tabular Scene. The grounding query is ``Cyprus''. The red bounding box in (a) is the ground-truth answer, while the blue one is the bounding box generated by our \Grounder\ module. The red bounding box in (b) is the output of Grounding DINO. }
	\label{fig:gazing_chart}
\end{figure*}


\begin{figure*}[h]
	\centering
	\begin{subfigure}{0.45\linewidth}
		\centering
		\includegraphics[width=1\linewidth]{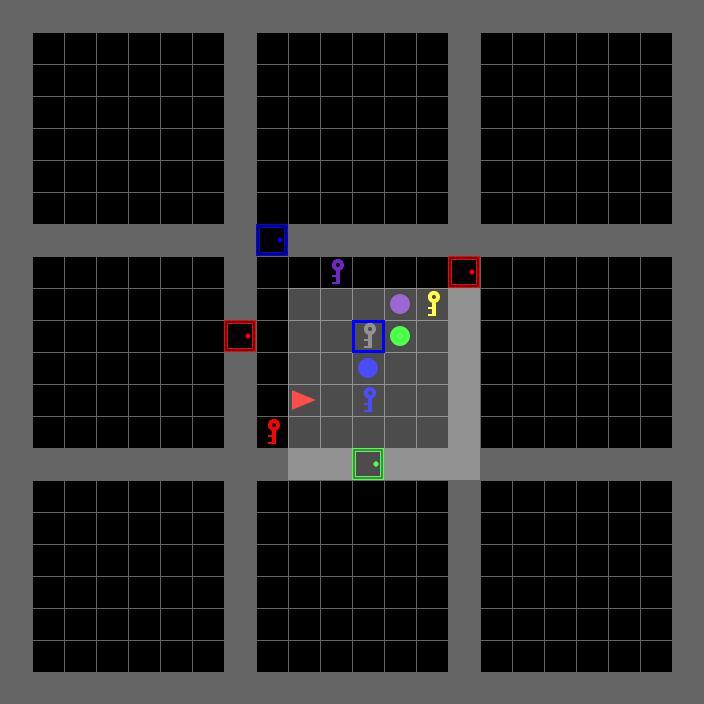}
		\caption{\Model\ }
		\label{subfig:Model_gazing_minigrid}
	\end{subfigure}
	\centering
	\begin{subfigure}{0.45\linewidth}
		\centering
		\includegraphics[width=1\linewidth]{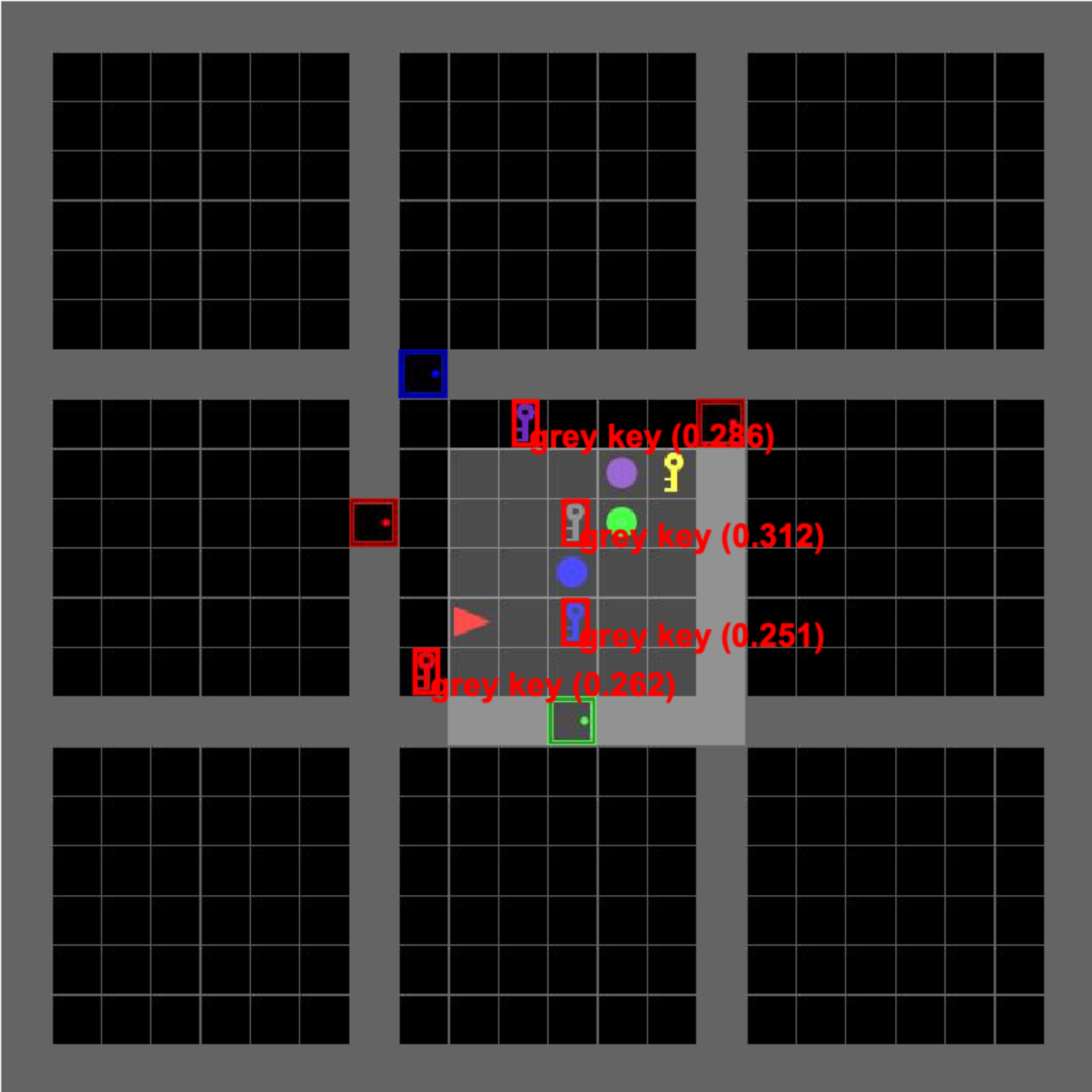}
		\caption{Grounding DINO}
		\label{subfig:DINO_gazing_minigrid}
	\end{subfigure}
	\centering
	\caption{The grounding comparison between Grounding DINO and the \Grounder\ module of \Model\ in Minigrid Scene. The grounding query is ``grey key''. The blue bounding box in (a) is generated by the \Grounder\ module of \Model, which overlaps the ground-truth red bounding box. Meanwhile, the red bounding box in (b) is the output of Grounding DINO. }
	\label{fig:gazing_minigrid}
\end{figure*}

\section{\Model\ }
\label{Appendix:model}

\begin{figure}[h]
  \centering
  \includegraphics[width=1\linewidth]{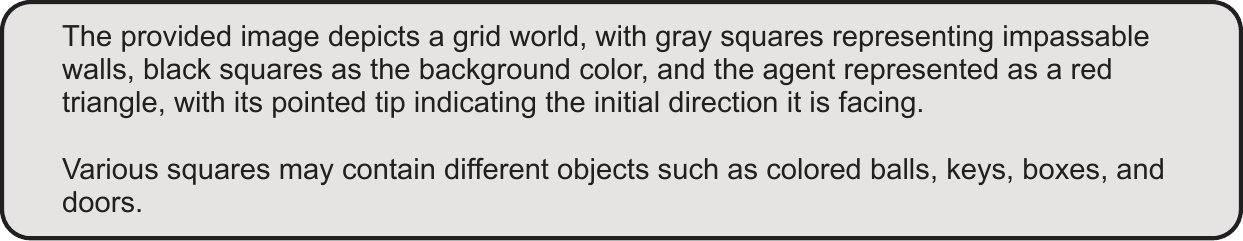}
   \caption{The description of Minigrid Scene added to the prompts. }
   \label{fig:prompt_babyai}
\end{figure}

Our \Model\ consists of two MLLMs, one for \Agent, and one for \Grounder. 
For each input question, \Model\ appends buffer information and queries to the respective prompts for \Agent\ and \Grounder. 
For images from Minigrid, a description of the Minigrid environment, as shown in \Cref{fig:prompt_babyai}, is included in both training and testing.
The remaining components of the \Agent\ prompt and \Grounder\ prompt are consistent across all three scenes. 

\begin{figure}[h]
  \centering
  \includegraphics[width=1\linewidth]{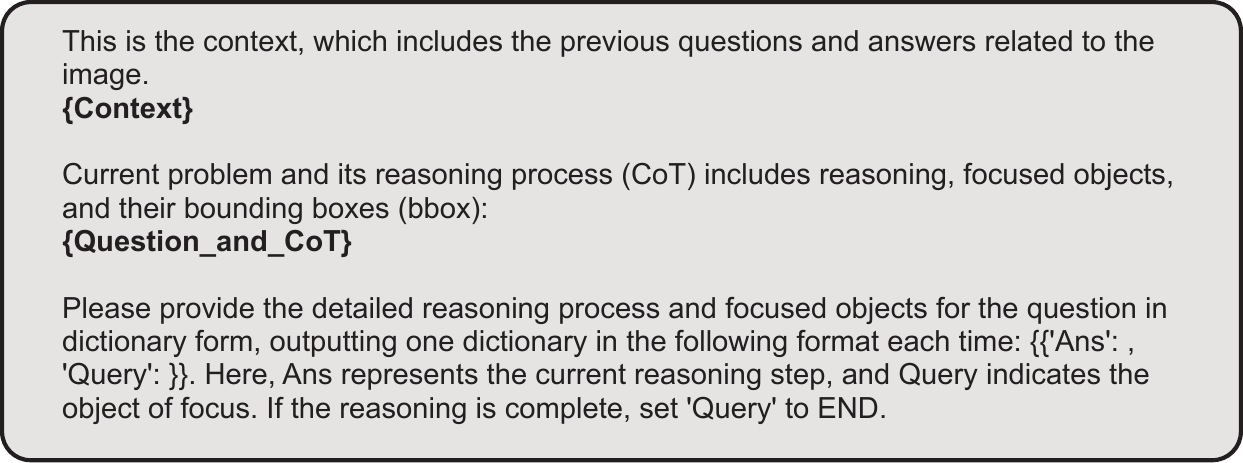}
   \caption{The prompt structure of the \Agent\ module when the last Query output of the \Agent\ module is not `END'. }
   \label{fig:prompt_cerebrate_0}
\end{figure}

\begin{figure}[h]
  \centering
  \includegraphics[width=1\linewidth]{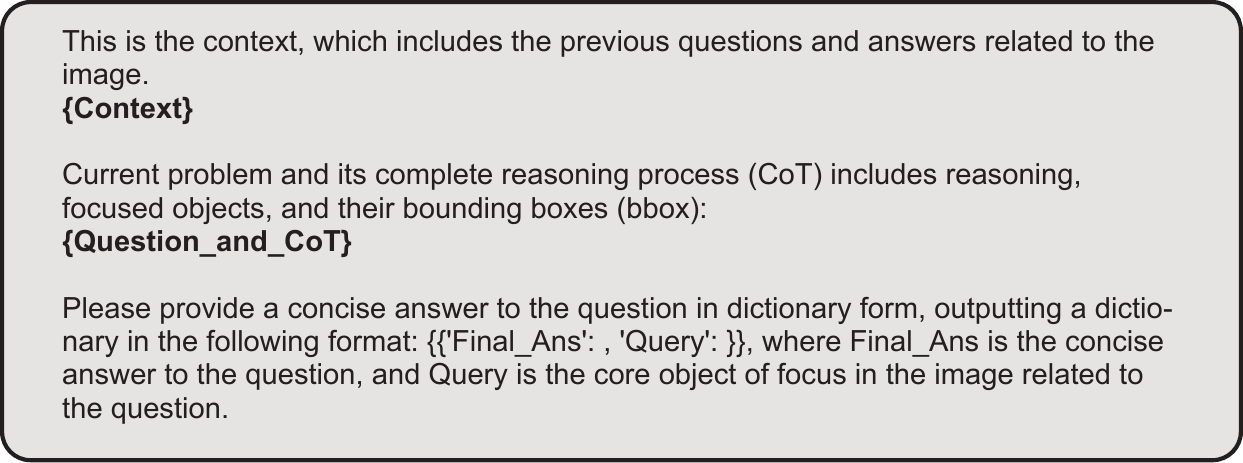}
   \caption{The prompt structure of the \Agent\ module when the last Query output of the \Agent\ module is `END'. }
   \label{fig:prompt_cerebrate_1}
\end{figure}

\noindent \textbf{\Agent\ Prompt. }
For deliberating, \Model\ provides the dialogue context and Chain of Thought (CoT) history for the current question in the prompt, as shown in \Cref{fig:prompt_cerebrate_0}.
When the `END' token appears in the latest `Query' from the \Agent\ module, signaling the end of the CoT process, \Model\ provides a new prompt, as shown in \Cref{fig:prompt_cerebrate_1}, to the \Agent\ module for generating the final answer.

\begin{figure}[ht]
  \centering
  \includegraphics[width=1\linewidth]{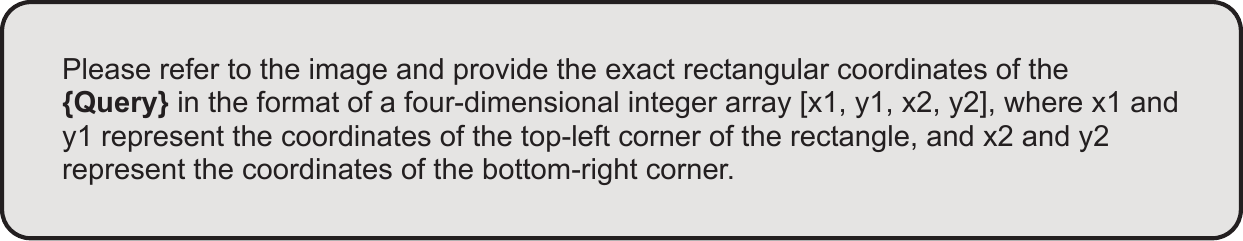}
   \caption{The prompt structure of the \Grounder\ module. }
   \label{fig:prompt_gaze_0}
\end{figure}

\noindent \textbf{\Grounder\ Prompt.}
For gazing, \Model\ extracts the `Query' from the output of the \Agent\ module and provides it to the \Grounder\ module along with the prompt shown in \Cref{fig:prompt_gaze_0}. 
The output from the \Grounder\ module, which includes the bounding box of the query, is then saved in the \Agent\ buffer to support the next turn of Deliberating.

\section{Implementation}
\label{Appendix:implementation}

The detailed parameters of implementation are shown in \Cref{tab:cerebration_param,tab:gazing_param}.

\section{Qualitative Comparison of Grounding}
\label{Appendix:qual_gaze}

\Cref{fig:gazing_chart,fig:gazing_minigrid} show a comparison of grounding ability between \Model\ and Grounding DINO~\cite{groundingdino}. 
As illustrated in \Cref{subfig:DINO_gazing_chart}, Grounding DINO struggles with grounding tasks involving Optical Character Recognition (OCR). 
In contrast, \Model\ leverages the generalization capability of LLMs, enabling it to effectively locate the target words, as shown in \Cref{subfig:Model_gazing_chart}. \Cref{subfig:DINO_gazing_minigrid} illustrates that Grounding DINO fails to handle objects with attributes. 
Although the grey key has a marginally higher confidence, accurately locating the `grey' key in the image confuses Grounding DINO.
In contrast, \Model\ accurately identifies the grey key in \Cref{subfig:Model_gazing_minigrid}, which aids the subsequent actions of the \Agent\ module. 

\section{Ablation Study}
\label{Appendix:ablation}

\begin{figure}[h]
  \centering
  \includegraphics[width=1\linewidth]{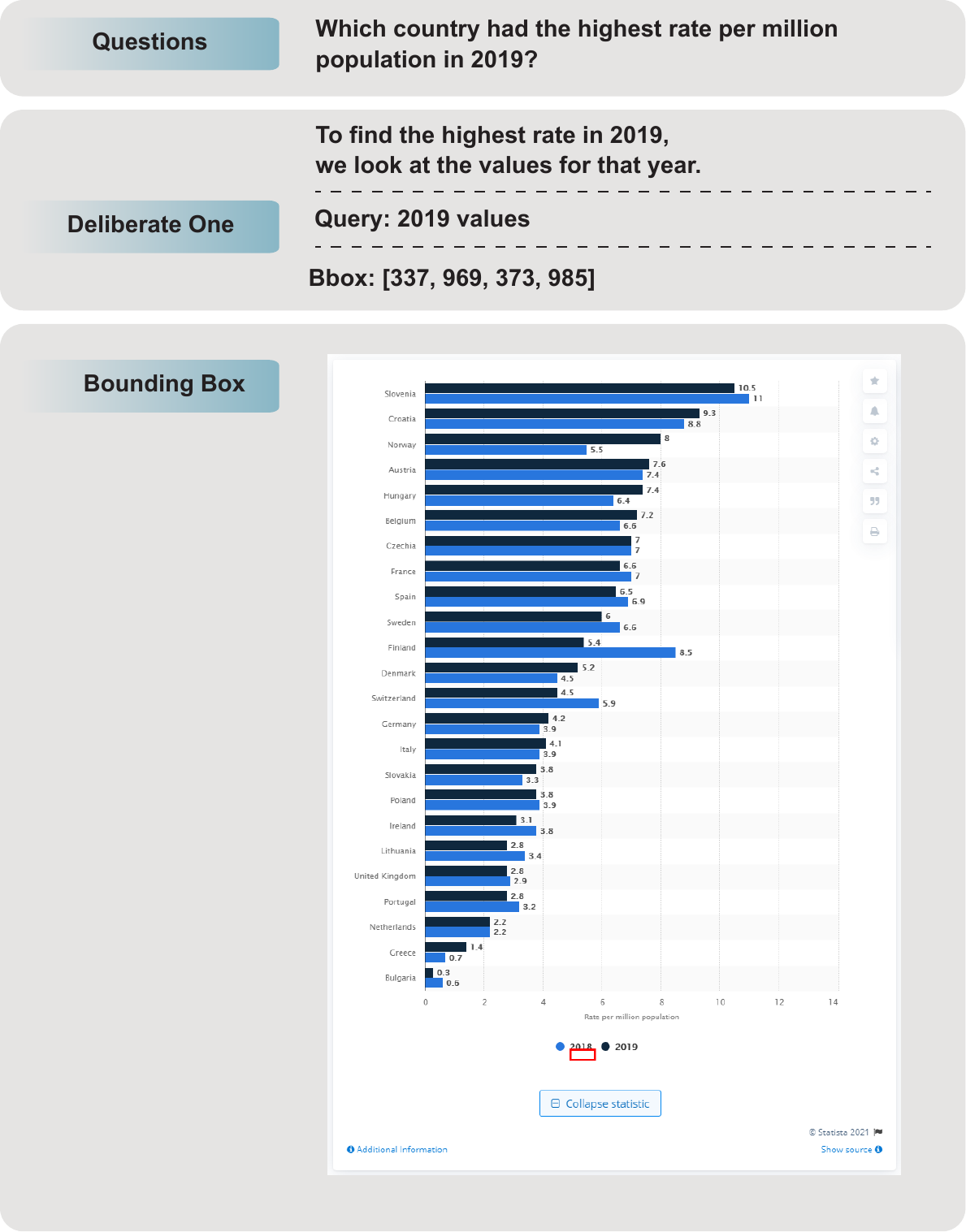}
   \caption{The second example of comparison between different MLLMs under everyday scenes.}
   \label{fig:chart_false}
\end{figure}

We observe a counterintuitive performance trend in Table~3 in the main paper: \Grounder\ provides only limited performance gains and, in some cases, even reduces performance, particularly in tabular and Minigrid scenarios. As shown in \Cref{fig:chart_false}, \Grounder\ incorrectly identifies the bounding box for a critical but tiny piece of information—the year 2019—misleading \Agent\ to focus on the wrong color bar. This issue accounts for most failure cases.  

To further analyze this, we evaluate the proportion of tiny key regions across different scenarios in \Dataset\ (\Cref{tab:tiny_experiment}). In tabular and Minigrid scenes, nearly all key regions occupy less than 3\% of the total image area, making them particularly challenging for \Grounder\ to detect accurately. To mitigate this, we curate an alternative test dataset for tabular scenes, excluding questions that require attention to extremely small regions. We then fine-tune Visual CoT and \Model\ with \Dataset\ and evaluate them on this revised tabular split. As shown in \Cref{tab:abla_experiment}, \Grounder's impact becomes more pronounced. \Cref{tab:general_data_experiment} demonstrates that \Model\ performs comparably or slightly lower on standard multimodal benchmarks, as it targets complex multi-region dialogues without in-domain training data.

\begin{table}[ht]
\centering
\scalebox{0.65}{
\begin{tabular}{c|c|c|cccc}
\toprule
Model &  Fine-tuning Data & \Grounder\ & T1 & T2 & T3 & T4\\
\midrule
Visual CoT-13B & \Dataset & - &2.00 & 1.43 & 0.40 & 0.95\\
\Model-14B  &  \Dataset & \ding{55} & 3.15 & 2.35 & 1.78 & 1.23 \\
\Model-14B  &  \Dataset & \ding{51} & 4.20 & 3.10 & 2.55 & 1.95 \\
\bottomrule
\end{tabular}
}
\caption{Tabular scenes results of MLLMs fine-tuned on MMDiag, using the same evaluation metrics as the previous evaluation. }
\label{tab:abla_experiment}
\end{table}

\begin{table}[!h]
\centering
\scalebox{0.65}{
\begin{tabular}{c|cccc}
\toprule[1pt]
Benchmark &  MMBench & MM-Vet & RefCOCO+ & RefCOCOg \\
\midrule
\Model-14B  & 63.7 & 28.5 & 0.834 & 0.775 \\
\bottomrule[1pt]
\end{tabular}
}
\caption{\Model\ performance on general datasets. }
\label{tab:general_data_experiment}
\end{table}

\begin{table}[!h]
\centering
\vspace{-2.5mm}
\scalebox{0.65}{
\begin{tabular}{c|ccccc}
\toprule[1pt]
Scenario &  $\leq 0.2\%$ &  $\leq 1\%$ &  $\leq 3\%$ & $\leq 5\%$ & $\leq 10\%$  \\
\midrule
Everyday & 7.57\% & 27.62\% & 47.99\% & 57.49\% & 69.91\% \\
Tabular  & 87.17\% & 99.24\% & 99.80\% & 99.92\% & 100\% \\
Minigrid  & 6.98\% & 66.61\% & 96.99\% & 99.41\% & 100\% \\
\bottomrule[1pt]
\end{tabular}
}
\caption{\Dataset\ tiny key regions percentage. }
\label{tab:tiny_experiment}
\end{table}

\section{Qualitative Comparison of Multi-Turn Multimodal Dialogue}
\label{Appendix:qual_cereb}

\begin{figure}[ht]
  \centering
  \includegraphics[width=1\linewidth]{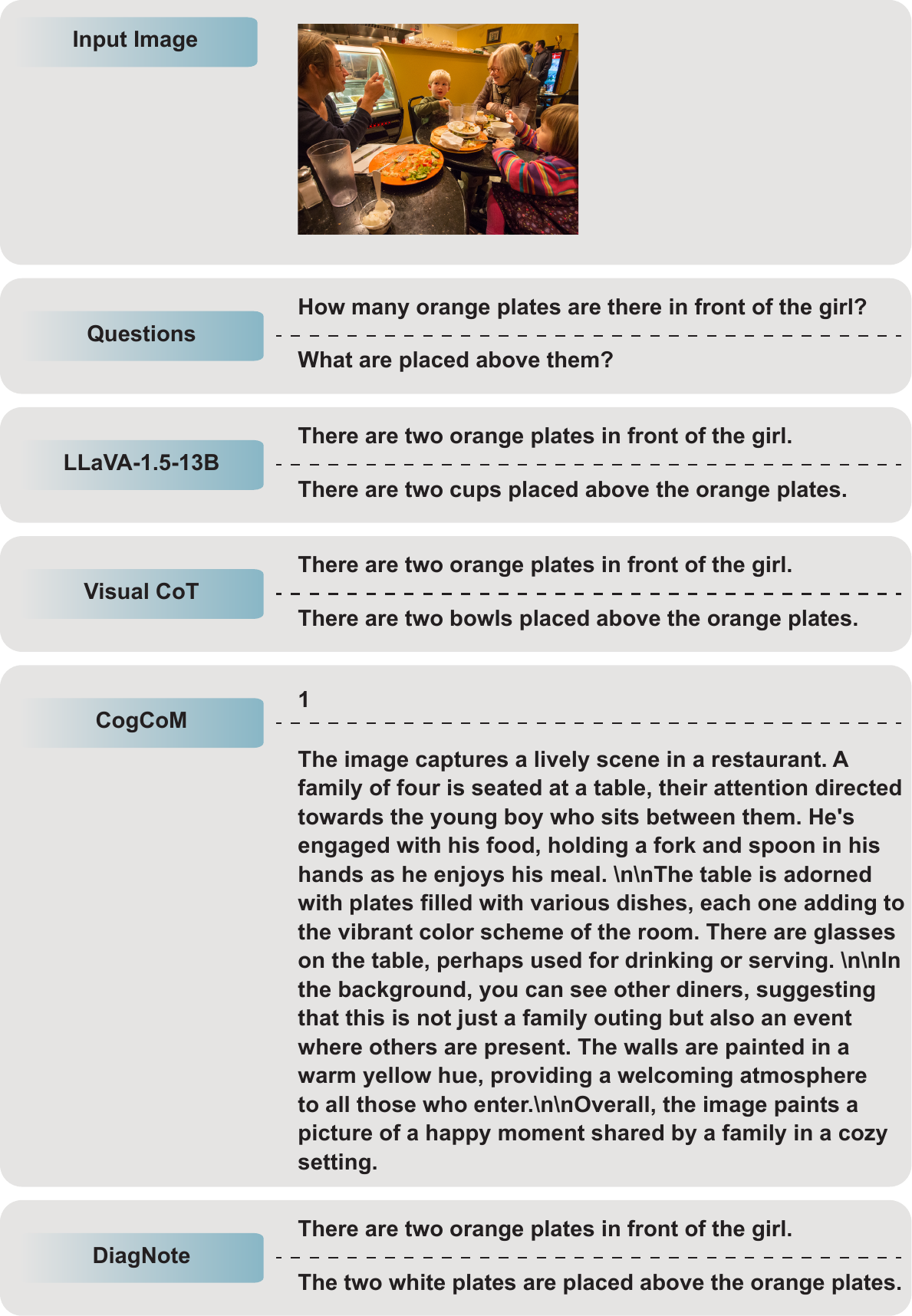}
   \caption{The first example of comparison between different MLLMs under everyday scenes.}
   \label{fig:everyday_case_1}
\end{figure}

\begin{figure}[h]
  \centering
  \includegraphics[width=1\linewidth]{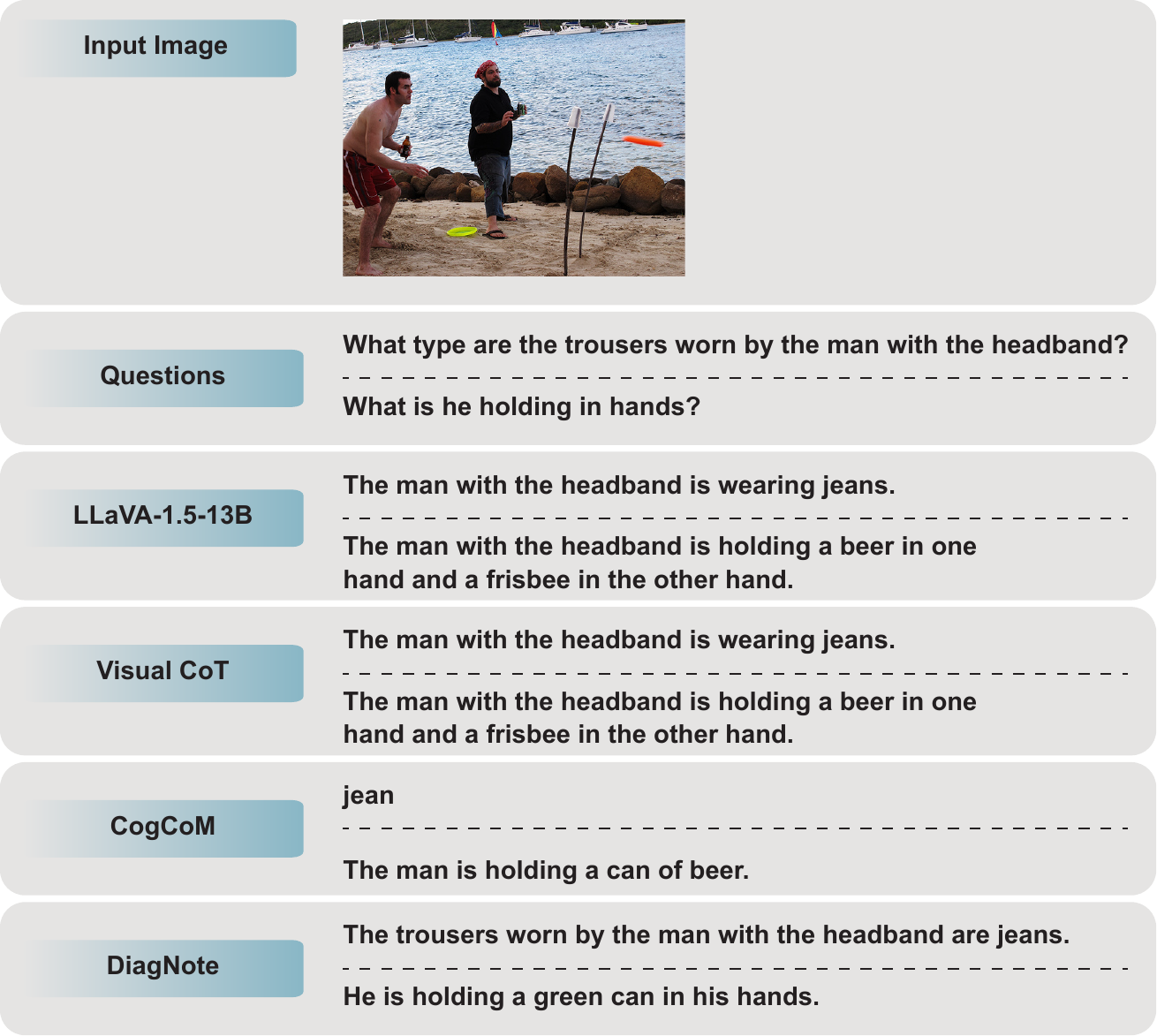}
   \caption{The second example of comparison between different MLLMs under everyday scenes.}
   \label{fig:everyday_case_2}
\end{figure}

We present several cases comparing models in everyday scenarios and tabular scenes.
\Cref{fig:everyday_case_1,fig:everyday_case_2} show examples from unseen everyday scenarios.
In \Cref{fig:everyday_case_1}, CogCoM~\cite{cogcom} completely fails to answer the two-turn questions correctly.
Despite the assistance of the counting expert, CogCoM is unable to answer the first counting question. 
Although LLaVA-1.5-13B~\cite{llava1.5} and Visual CoT~\cite{viscot} can answer the first questions accurately, both encounter hallucinations when responding to the second question, mistakenly identifying white plates as cups and bowls, respectively.
In contrast, our \Model\ performs well on both questions, demonstrating the effectiveness of the \Grounder\ module in ensuring \Model\ stays grounded in visual details.
In \Cref{fig:everyday_case_2}, CogCoM fails to provide a clear answer to the first question, instead offering a confusing single word `jean'.
Again, LLaVA-1.5-13B and Visual CoT answer the first question correctly, but imagine the man was holding a frisbee.
Both CogCoM and \Model\ understand the context, with \Model\ accurately describing the can based on the visual details.
In contrast, CogCoM mistakenly assumes it is a can of beer, which may not be the case.

\begin{figure}[H]
  \centering
  \includegraphics[width=1\linewidth]{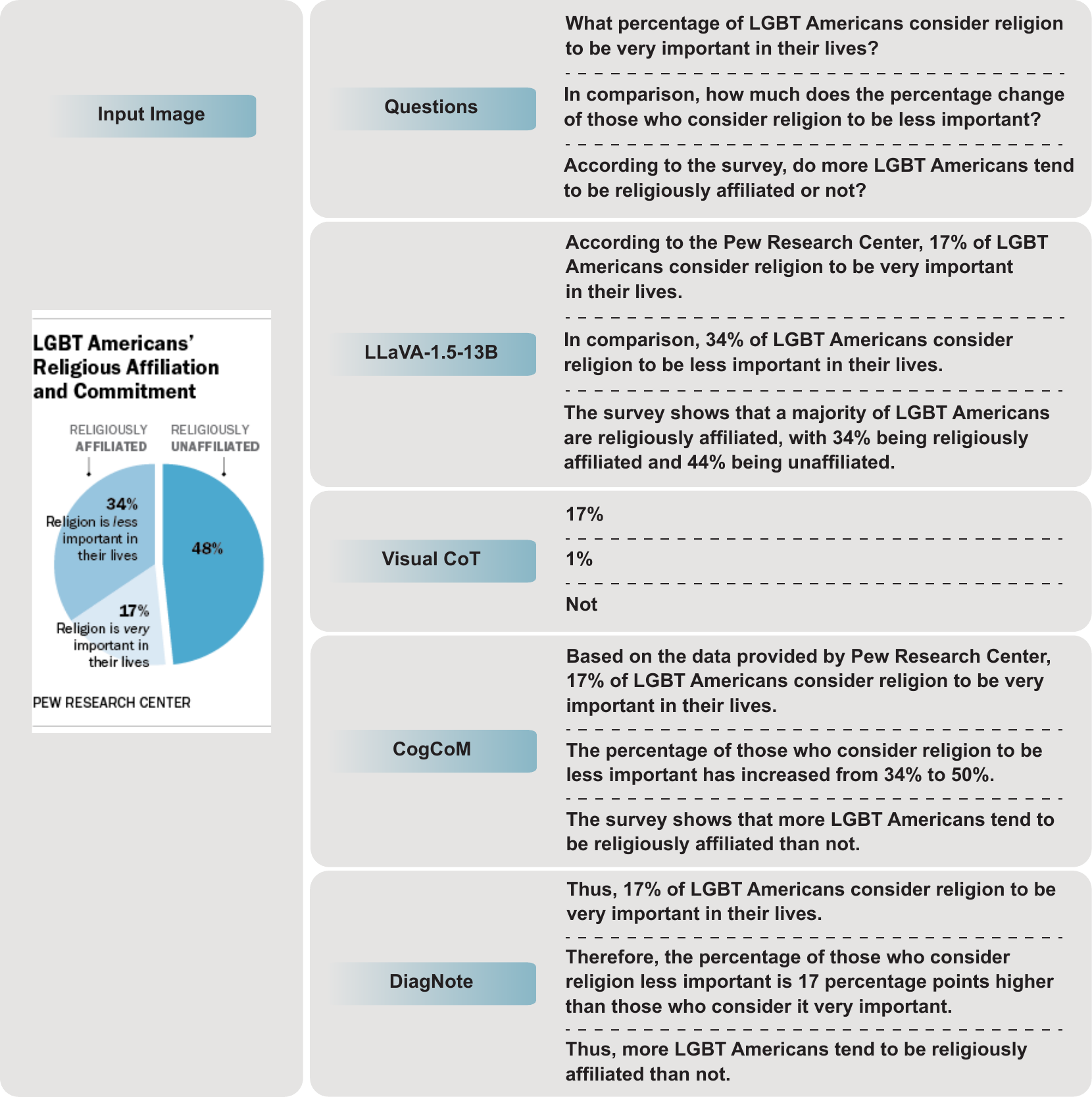}
   \caption{One example of comparison between different MLLMs under tabular scenes.}
   \label{fig:tabular_case_1}
\end{figure}

\Cref{fig:tabular_case_1} presents examples of unseen tabular scenes.
All models answer the first question correctly. 
However, Visual CoT provides a completely incorrect answer to the second question, while CogCoM introduces an unfounded `50\%'. 
LLaVA-1.5-13B correctly identifies the visual detail `34\%', but overlooks the keyword `change' in the question, which requires a calculation between two percentages.
Only \Model\ answers the question precisely.
The final question requires the models to understand the entire pie chart. 
The model should compare the sum of two parts on the right side of the pie chart with the left part to obtain the final answer `yes'.
Visual CoT fails to provide this correct answer, and LLaVA-1.5-13B misinterprets the unaffiliated percentage and derives an incorrect affiliated percentage.
Both CogCoM and \Model\ reach the right conclusion.
Overall, \Model\ performs well on all questions, demonstrating its ability to focus on both visual and language details and to comprehend the full picture the chart conveys. 
This strong ability can be attributed to the \Grounder\ and \Agent\ structure, which enables it to zoom in on specific details while integrating multimodal information for a holistic understanding.

\end{document}